\documentclass[journal]{IEEEtran}
\usepackage{multirow}
\usepackage{graphicx}
\usepackage{amsmath}
\usepackage{makecell, longtable}
\usepackage{nccmath}
\usepackage{enumitem}
\usepackage{url}
\usepackage[hidelinks]{hyperref}
\usepackage [font=footnotesize]{caption}
\usepackage[table, dvipsnames]{xcolor}

\usepackage{tabularx}
\usepackage{threeparttable}
\usepackage{amssymb}
\usepackage{threeparttable}
\usepackage{booktabs}
\usepackage{float}
\definecolor{maroon}{cmyk}{0,0.87,0.68,0.32}
\ifCLASSINFOpdf
\else
 
\fi

\begin{document}
%
\title{Safety Implications of Explainable Artificial Intelligence in End-to-End Autonomous Driving}
%
%
%

\author {Shahin Atakishiyev, Mohammad Salameh, Randy Goebel \thanks{Shahin Atakishiyev is a Postdoctoral Fellow in the Department of Computing Science at the University of Alberta, Edmonton, AB, Canada. Email address: shahin.atakishiyev@ualberta.ca
}
\thanks{Mohammad Salameh is a Principal Researcher at Huawei Technologies Canada, Edmonton, AB, Canada. Email: mohammad.salameh@huawei.com}
\thanks{Randy Goebel is a Fellow of the Alberta Machine Intelligence Institute, and Professor in the Department of Computing Science, University of Alberta, Edmonton, AB, Canada. Email: rgoebel@ualberta.ca}
\thanks{Manuscript was accepted for publication on May 25, 2025.}
}

%
%

\markboth{}%
{Shell \MakeLowercase{\textit{et al.}}: Bare Demo of IEEEtran.cls for IEEE Journals}
%



\maketitle

\begin{abstract}
The end-to-end learning pipeline is gradually creating a paradigm shift in the ongoing development of highly autonomous vehicles (AVs), largely due to advances in deep learning, the availability of large-scale training datasets, and improvements in integrated sensor devices. However, a lack of explainability in real-time decisions with contemporary learning methods impedes user trust and attenuates the widespread deployment and commercialization of such vehicles. Moreover, the issue is exacerbated when these vehicles are involved in or cause traffic accidents. Consequently, explainability in end-to-end autonomous driving is essential to build trust in vehicular automation. With that said, automotive researchers have not yet rigorously explored safety benefits and consequences of explanations in end-to-end autonomous driving. This paper aims to bridge the gaps between these topics and seeks to answer the following research question: What are safety implications of explanations in end-to-end autonomous driving? In this regard, we first revisit established safety and explainability concepts in end-to-end driving. Furthermore, we present critical case studies and show the pivotal role of explanations in enhancing driving safety. Finally, we describe insights from empirical studies and reveal potential value, limitations, and caveats of practical explainable AI methods with respect to their potential impacts on safety of end-to-end driving.
\end{abstract}


\IEEEpeerreviewmaketitle
\section{Introduction}

\IEEEPARstart {O}{ver} the last two decades, interest in developing operationally safe and reliable AVs has gained significant momentum. With the promises of reducing traffic accidents, enhancing safety \cite{hicks2018safety} and environmental sustainability \cite{liu2019can}, the field of autonomous driving has received considerable attention from both industry and academic institutions. The Society of Automotive Engineers (SAE) has defined a classification system to assess the level of autonomy in AVs \cite{Shuttleworth}. According to this classification, the autonomous decision-making ability of these vehicles ranges from Level 0 (no driving automation) to Level 5 (no human intervention in any traffic situation). Starting from Level 3, most human-controlled functions of a vehicle are replaced with autonomous control methods, resulting in high automation.

\begin{figure}[htp]
    \centering
    \includegraphics[width=8.8 cm]{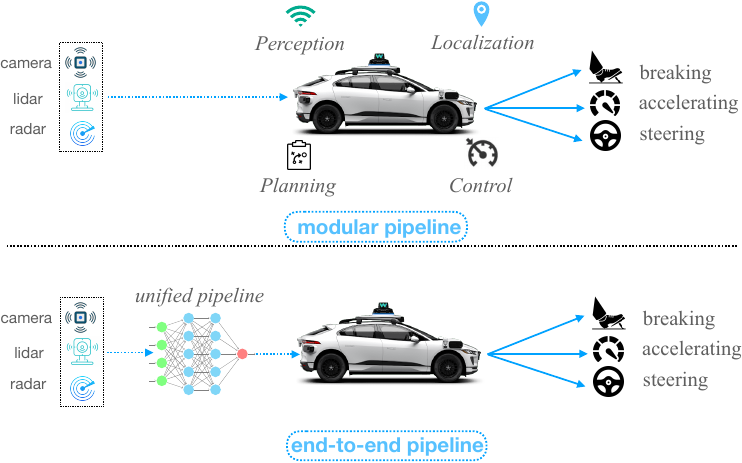}
    \caption{The modular and end-to-end pipeline comparison in autonomous driving. The modular pipeline is more explainable by design as it contains several interlinked components, but an error in one module (e.g., misdetection in perception) cascades to the downstream module (planning/control). On the other hand, the end-to-end driving pipeline, while lacking explainability, directly maps sensor input to control commands, optimizing system performance. Continuous improvement through data-driven optimization makes end-to-end learning more scalable. The source of the image of the vehicle: Waymo.}
    \label{fig:e2e}
\end{figure}
\hspace{-0.47cm}
There are two primary design architectures for autonomous driving systems: modular and end-to-end pipelines. The modular pipeline consists of interlinked components such as perception, localization, planning, and control and has been a conventional way of building autonomous driving systems \cite{tampuu2020survey, yurtsever2020survey}. In the modular architecture, an AV uses a variety of sensor suites, such as camera, LIDAR, radar, etc., and perceives its environment (perception), locates itself on the road via GPS, Inertial Measurement Unit (IMU) data (localization), plans a safe path based on the perception and localization data (planning), and executes the planned trajectory by sending commands to the vehicle's actuators, such as steering, throttle, and brakes (control). The advantage of such a design is that individual modules make the system more traceable and debuggable, and the origin of errors in the holistic system can be identified easily. On the other hand, such errors are propagated forward to the subsequent layers and become cumulative, affecting the safety and effectiveness of the modular system adversely \cite{hu2023planning, chen2023end}. \\
To address the aforementioned issue, the end-to-end learning approach has recently been proposed as a paradigm shift in the ongoing development of autonomous driving technology \cite{chen2023end, bojarski2016end, coelho2022review}. End-to-end learning directly inputs sensory information and maps it to control commands as a unique machine learning (ML) task \cite{chen2023end, chib2023recent}. Relying on a unified model without explicit, individual components hinders the explainability aspect of end-to-end learning as a downside. On the other hand, this learning paradigm has two primary advantages - \textit{ potential safety enhancement} and \textit{computational efficiency} benefits over its counterpart. End-to-end driving is not \textit{universally} safer than the modular pipeline; however, the automotive industry argues that this pipeline could eventually become safer in the real world as it promises reduced error propagation, better handling of uncertainty, and lower processing latency \cite{chen2023end, chib2023recent}. Furthermore, the computational efficiency, in terms of elimination of intermediate representations, parallelization (GPUs/TPUs for parallel tensor operations), and streamlined data processing are improved substantially due to shared backbones \cite{chen2023end, chib2023recent}. Hence, big automotive companies have put tremendous efforts into leveraging the end-to-end driving approach for their AVs, thanks to this pipeline's enhanced safety and efficiency promises. Waymo’s EMMA \cite{hwang2024emma}, Wayve’s AI Driver \cite{wayve_ai_driver}, NIO’s Banyan \cite{nio2024e2e}, XPENG’s XNet \cite{xpeng2024ai}, and Li Auto’s OneModel \cite{liAuto2024onemodel} are examples of recent industrial end-to-end autonomous driving models. Meanwhile, recent attempts to explain the behavior of deep neural networks \cite{samek2021explaining} have been a motivation for automotive researchers to present intelligible information to target groups so that relevant people understand the behavior of the black-box system in end-to-end driving. \\
Explanations have recently been explored for various intelligent transportation systems and autonomous driving, primarily with the goal of increasing public trust, system transparency, and improving situational awareness \cite{schmidt2021can, kim2023and, atakishiyev2021explainable, gyevnar2024, adadi2023explainable, chen2024maritime}. However, we argue that the value of explanations goes beyond these measures, and explanations can also contribute to the safety analysis of AVs. Remarkably, a recent study has provided valuable insights into the role of explanations for safe and trustworthy AVs with a conceptual framework for modular autonomous driving \cite{kuznietsov2024explainable}. Meanwhile, to the best of our knowledge, automotive researchers have not thoroughly explored the safety benefits and consequences of explanations in end-to-end driving, and the literature remains scarce with relevant studies. Consequently, to bridge the gaps between these topics, this paper is explicitly centered on understanding safety benefits and consequences of XAI approaches in end-to-end autonomous driving by answering the following research question: \textit{What are potential safety implications (benefits and ramifications) of explanations in end-to-end autonomous driving?}
\vspace{0.35cm}
\subsection{Main Contributions}
Focusing on answering the question outlined above, the key contributions of our paper are threefold:

\begin{itemize}[leftmargin=*]
\item We present case studies to demonstrate how explaining temporal decisions, both in real-time and retrospectively, can enhance safety of end-to-end driving;
\item We draw insights from empirical studies and describe opportunities, challenges, and limitations of XAI approaches with respect to their impacts on safety of end-to-end driving;
\item We describe the unique benefits of explanations for end-to-end driving compared to modular driving by revealing insights from the conducted analytical and empirical studies. 
\end{itemize}
\begin{figure}[htp]
    \centering
    \includegraphics[width=8.8 cm]{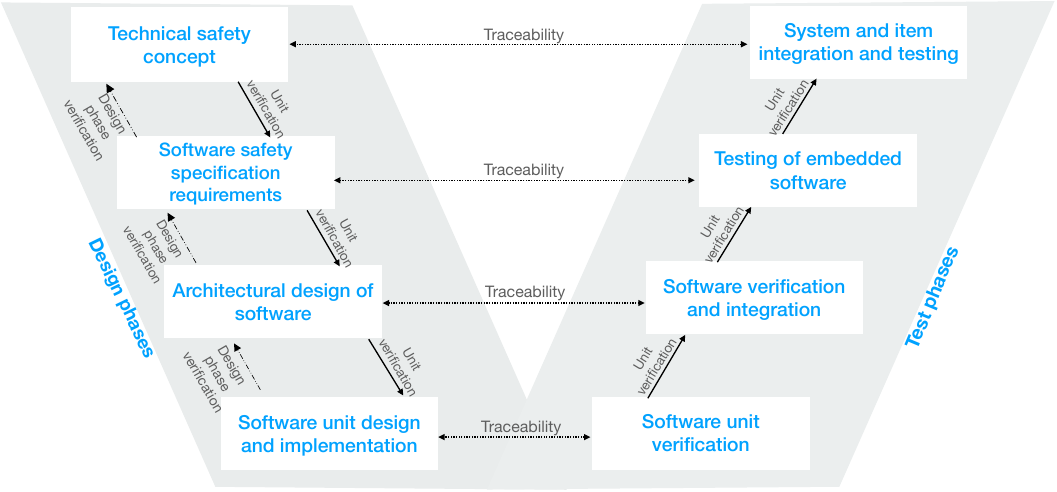}
    \caption{V-model in ISO 26262. The figure drawn based on the content in \cite{V_model}.}
    \label{fig:V-model}
\end{figure}

\subsection{The Structure of the Paper}
Our paper is organized as follows. Following the introduction, a brief overview of established safety concepts, learning paradigms, and XAI approaches in end-to-end driving are presented as background information in Sections II, III, and IV, respectively. Afterward, the main contributions of the paper are presented: Section V provides analytical studies where XAI approaches can bring an enhanced safety benefit to end-to-end driving. Section VI presents insights from experimental studies on the explainability-informed safety assessment proposition. Future prospects on underexplored XAI concepts and their potential safety ramifications in end-to-end driving are covered in Section VII. Finally, Section VIII concludes the paper with key findings and an overall summary.

\section{Safety of autonomous driving}
The safety of the deployed software system in autonomous driving is a crucial factor for building confidence in this technology. As ML approaches play an indispensable role in enabling an AV to make intelligent decisions, these approaches need to be safety-compliant both theoretically and practically. ISO 21448, also known as the "Safety of the Intended Functionality" (SOTIF), is an international standard that provides guidelines to ensure the safety of road vehicles, particularly those relying on complex sensors and algorithms \cite{ISO21448}. It complements ISO 26262 \cite{iso201126262}, which has been specifically tailored to ensure the safety of electrical and electronic systems within passenger vehicles. However, explicit guidelines for AI functionalities in AVs are missing. Consequently, as a further complement,  ISO/PAS 8800 \cite{ISO8800}, a more recent standard, is under development for the safety of AI technology in AVs, explicitly focusing on the potential risks of unintended safety-related behaviors in vehicles due to model outputs, systematic faults and random hardware failures in the AI components within the AV, thus urging hardware reliability. Furthermore, with increasing reliance on real-time perception data, AI, and vehicular communication technologies, today’s AVs are more prone to cyberattacks, necessitating compliance with the principles of the ISO/SAE 21434 standard \cite{ISO_cybersecurity}.  Moreover, AVs may require takeover by backup drivers due to potential unexpected scenarios such as adverse weather conditions, approached construction zones, missed lane boundaries \cite{mcdonald2019toward}, and related issues in terms of software, hardware failures, and cybersecurity issues, covered above (see Figure \ref{fig:take_over}), where this case is typically applicable to driving automation below Level 4 \cite{Shuttleworth, SAEJ3016_2021}. The transition of control from the intelligent driving system of an AV to the human operator happens in a short time interval and consists of two primary steps: 1) an AV makes a takeover request (TOR), and a human driver should receive this request immediately and take over the control of the steering wheel and pedals, and 2) the post-takeover step, where the driver takes over the control of the AV and manually performs the decisive action safely as per the traffic scenario \cite{huang2022effects}. The time granularity of these steps may vary from situation to situation, but the overall length of takeover cases is usually a few seconds. Consequently, a human actor must dominate takeover situations in the allotted time interval; otherwise, collision or other serious consequences may become inevitable. Finally,  while the transition to highly autonomous driving is a significant change in the intelligent capabilities of AVs, the immense need to operate safely in possible failure cases also escalates. Fail-safe mechanisms often include redundancy, where backup systems take over if the primary system fails. In the context of autonomous driving,  a fail-safe is the ability of an AV to transition to a safe state, such as slowing down, pulling over, or stopping entirely in case of failure of the driving system due to various reasons at specific moments \cite{pek2019ensuring, vom2020fail}. Overall, the established definition of safe autonomous driving requires risk minimization and assurance in seven key tasks, namely, pedestrian detection, drowsy driver detection, vehicle detection, road detection, lane detection, traffic sign detection, and collision avoidance, as identified by \cite{muhammad2020deep}, and end-to-end driving, in turn, must also pass these crucial safety checks.

\section{End-to-End Learning Paradigms}
This section presents some fundamental approaches behind end-to-end autonomous driving. In this context, the following subsections describe a brief overview of three leading approaches: 1) reinforcement learning, 2) imitation learning, and  3) a recently emerging technique, differentiable learning, enabling end-to-end driving for AVs.  
\subsection{Reinforcement Learning}
The operational environment of AVs is dynamic and includes consideration for a wide variety of traffic circumstances. In all such scenarios, it is indispensable that the intelligent driving system senses the scene accurately and
maps the perceived information to appropriate actions. Reinforcement Learning (RL) is a computationally robust approach to solve such sequential decision-making problems and its theory is formalized as a Markov Decision Process (MDP) \cite{sutton2018reinforcement}, comprising 
  $S$: a set of a \textit{state} space of an environment;
     $\mathcal{A}$: set of \textit{actions} the agent can select from;
     $T$: a \textit{transition probability} $T(s_{t+1}| s_t, a_t)$, denoting the probability that the environment will transition to state $s_{t+1}\in S$ if the agent takes action $a\in \mathcal{A}$ in state $s \in S$;
     $R$:  a \textit{reward} function where $r_{t+1} = R(s_t, s_{t+1})$ is a reward acquired for taking action $a_t$ at state $s_t$ and transitioning to the next state $s_{t+1}$; and
     $\gamma$: a discounting factor, defined as $\gamma \in (0, 1] $.
An agent's behavior in its environment

\begin{figure}[H]
    \centering
    \includegraphics[width=8.8cm]{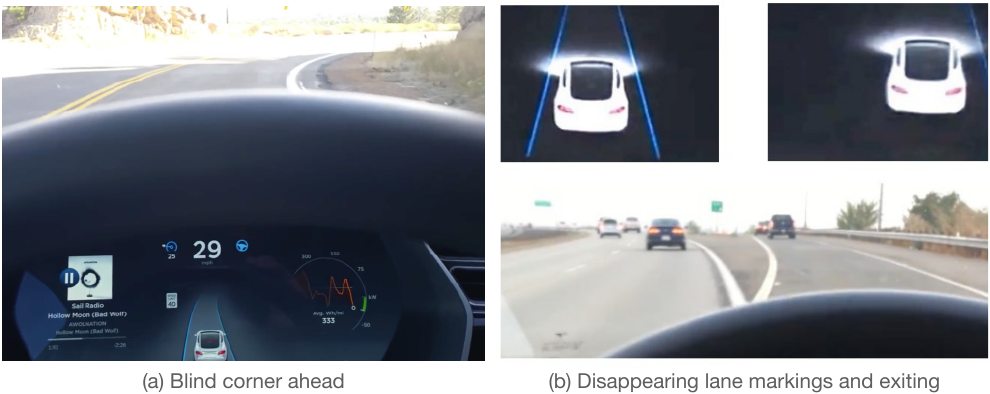}
    \caption{Potential takeover situations: (a) The blind corner ahead reduces the AV's perception ability and urges for a takeover of the human driver, and (b) Autopilot perceives edge-markings of an exit lane as the continuation of the present lane (upper left, while still on the main road), steers right and the vehicle exits the road incorrectly (upper right, exit). Edge markings disappear when the autopilot takes a wrong exit. Images adapted from \cite{brown2017trouble}.}
    \label{fig:take_over}
\end{figure} \hspace{-0.49cm}
is defined by a rule, called a \textit{policy} $\pi$, which maps states
to a probability distribution over the actions $ \pi \colon
\mathcal{S} \to \mathcal{T(A)} $. The process can be modeled as an MDP with
a state space $\mathcal{S}$, action space $\mathcal{A}$, an initial state distribution $p(s_1)$, transition dynamics $p(s_{t+1} | s_t,
a_t)$, and reward function $r(s_t, a_t)$. At any state, the sum of the discounted future reward constitutes the return for that state:

\begin{equation}
R_t = \sum_{i = t}^{T} \gamma^{(i - t)} r(s_i, a_i) 
\end{equation}
With this framework, the goal of an RL agent is to find an optimal policy that maximizes the return from the starting distribution:

\begin{equation}
  J = \max_{\pi} \mathbb{E}_{r_i, s_i \sim E, a_i \sim \pi}\left[ R_1 \right]
\end{equation}
Given the agent follows a policy $\pi$ by taking an action $a_t$ in state $s_t$, the action-value (also known as Q-value) function, describing its expected return is defined as follows:

\begin{equation}
  Q^\pi(s_t, a_t) = \mathbb{E}_{r_{i \ge t}, s_{i > t} \sim E, a_{i > t} \sim \pi} \left[ R_t | s_t, a_t \right ]
\end{equation}
Adapting RL to autonomous driving, an AV can be thought of as an \textit{agent}, its immediate position as a \textit{state}, the driving scenario as an \textit{environment}, and depending on how correctly it chooses real-time action, the relevant metric as a scalar \textit{reward} approximating correctness of decision-making.\\
\subsection{Imitation Learning}
Initiation learning (IL) attempts to mimic the behavior of a human expert \cite{hussein2017imitation, pan2020imitation}. An IL agent imitates the behavior of an expert driver and learns from the driver's actions via demonstrations. Within this setting, the goal of the IL agent is to match its learning policy to the one demonstrated by an expert. There are two well-known categories of IL:
\subsubsection{Behavior cloning (BC)} In this category, given the dataset collected from an expert's behavior, the agent aims to match the expert's policy by minimizing the planning loss with respect to that dataset. Overall, BC is a simple and computationally efficient approach as it relies on huge amount of data provided by a human expert and works well for simple driving tasks \cite{tampuu2020survey, pomerleau1988alvinn}. However, complex driving situations are challenging for BC \cite{codevilla2019exploring}. 
\begin{figure*}[htp!]
 \centering
    \includegraphics[width=17 cm]{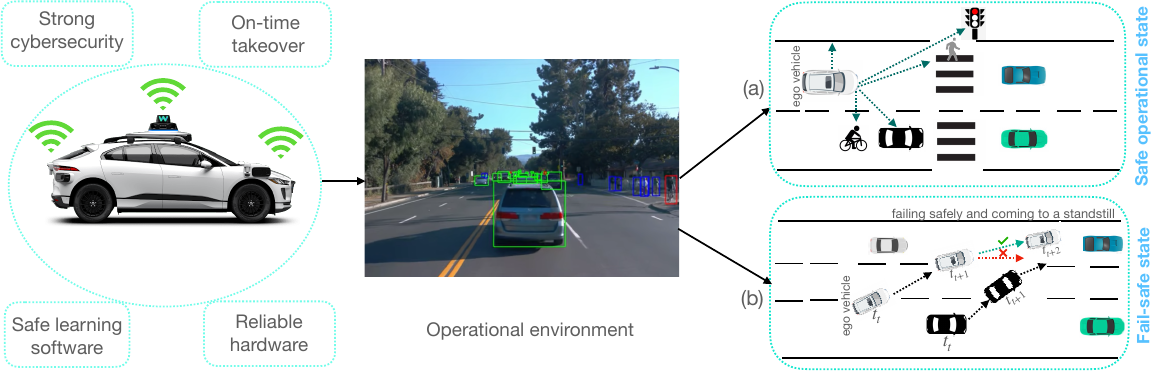}
    \caption{A diagram of safe autonomous driving. In (a), an AV (i.e., ego vehicle) interacts with the dynamic and stationary objects in the environment (i.e., vehicles ahead, pedestrians, and cyclists at the roadside identified with green, blue, and red bounding boxes, respectively) safely and keeps a distance from them. In (b), the ego vehicle faces the unexpected action of the other vehicle, understands its limited motion ability at that moment, and comes to a standstill as it can not drive safely at that time step.}
    \label{fig:safe_AD}
\end{figure*} 
\subsubsection{Inverse Reinforcement Learning (IRL)} Also known as Inverse Optimal Control, IRL aims to learn the expert's reward function. Once the reward function is inferred, that policy is further improved with RL. IRL is computationally more complex than BC; however, in terms of generalization in rare and unseen states, IRL has been effective across several self-driving tasks \cite{abbeel2004apprenticeship, phan2023driveirl}. 
\subsection{End-to-End Differentiable Learning}
This concept is a new approach to end-to-end autonomous driving. Built with a focus on planning-oriented driving \cite{hu2023planning, casas2021mp3, karkus2023diffstack, lu2024activead}, and also known as modular end-to-end planning, this learning methodology optimizes several modules while planning is prioritized as a goal of the hybrid architecture. The architecture of differentiable learning allows gradients to propagate seamlessly through all components during training, enabling optimization via backpropagation \cite{liu2024integrating}. Thus, the differentiable aspect ensures compatibility between learning-based modules and optimization processes, enabling seamless driving task alignment in a single, trainable model \cite{liu2024integrating, zhang2025bridging}. In this way, prediction and control tasks are jointly optimized while the unified system maintains its end-to-end trainability. Overall, the literature is not rich yet with such a learning pipeline as this trend is emergent, particularly with a keen interest in the automotive industry \cite{chen2023end}, given that both modular and end-to-end pipeline individually have their inherent benefits and their combinations can pave the way for the future of autonomous driving.

\section{XAI in end-to-end autonomous driving}
Incorporating explainability techniques into end-to-end autonomous driving has recently received increasing attention from automotive researchers. Overall, these techniques involve insights from both algorithmic methods regarding content and form as well as from time granularity and faithfulness perspectives of explanations. In this sense, this section presents XAI approaches from an algorithmic point of view and discusses the timing mechanism and credibility of explanations for enhancing safety of end-to-end driving.   
\subsection{Visual Explanations for End-to-End Autonomous Driving}
Visual explanations are usually \textit{post-hoc} or \textit{intrinsic} and classified as follows \cite{zablocki2022explainability}: 
\subsubsection{{Post-hoc Explanations}}
Post-hoc explanations describe actions retrospectively based on visual data from the video camera and other sensors. These explanations can be \textit{local} and \textit{global} depending on their task coverage. Local explanatory techniques justify the output of a particular task in the predictive model. On the other hand, global explanations describe the decisions of the entire predictive model. We summarize these techniques separately, below.

\textit{(1.1) {Local Explanations:}} There are two types of vision-based local explanations: (a) \textit{saliency methods}, which shows which components of an image have more influence on the predictive output, and (b) \textit{counterfactual explanations} that intend to find a causal relationship in image components that influence the predictive model's outcome.

\textit{(1.1.1) {Saliency methods:}}
Bojarski et al. \cite{bojarski2016end}'s study was a preliminary work that used a CNN method to map raw pixels to steering command and achieved impressive results, and subsequent work focused on the interpretability aspect of such end-to-end learning \cite{bojarski2017explaining, bojarski2018visualbackprop}.
 Some other notable research has also leveraged post-hoc saliency methods; these works include a causal attention mechanism \cite{kim2017interpretable, kim2021toward}, driving affordances \cite{sauer2018conditional}, and driving behavior understanding \cite{liu2020interpretable}.

\begin{figure}[t!]
\includegraphics[width = \columnwidth]{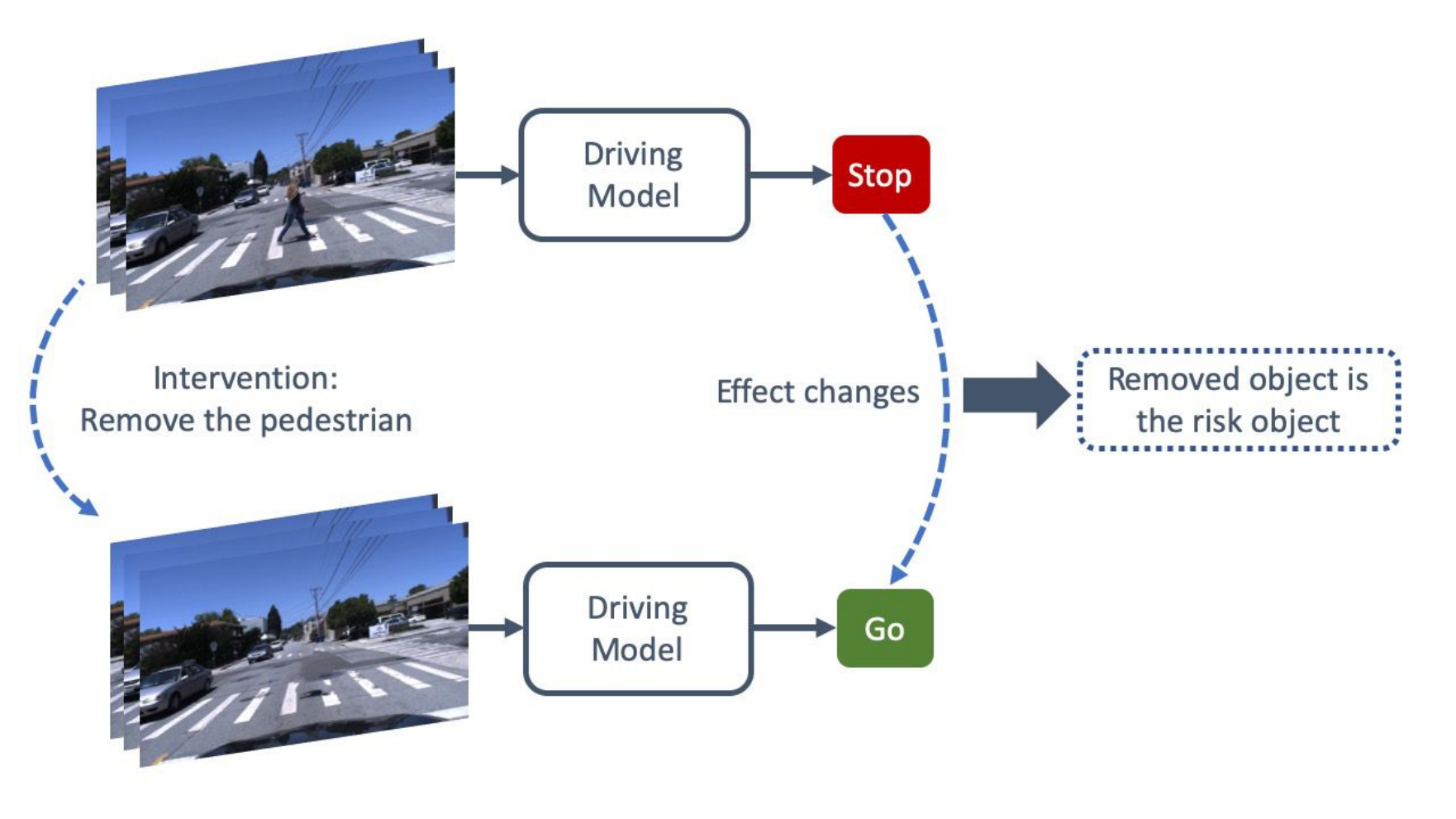}
\vspace{-18pt}
\caption{A causal effect of removing a pedestrian from a scene: The driving behavior changes from "Stop" to "Go" by showing that the eliminated object is a risk object for the "Stop" command. Graphics credit: \cite{li2020make} }
\label{fig:command_change}
\end{figure}

\textit{(1.1.2) {Counterfactual Explanations:}}
These explanations aim to identify causal relationships between different events by considering how the outcome of a model would have been different if the provided input had been modified in some way. In autonomous driving, the following can be a typical example of a counterfactual explanation: “Given the traffic scene, how can it be modified so that the AV turns right instead of driving straight?” In this case, the decision made by the AV becomes different from its actual prediction. We can also say that counterfactual explanations can be used to assess the potential consequences of different actions or decisions made by an AV. Examples of these explanation techniques are ChauffeurNet, \cite{bansal2019chauffeurnet}, its augmented version by Li et al. \cite {li2020make}(see Figure \ref{fig:command_change}), STEEX \cite{jacob2022steex}, and OCTET \cite{zemni2023octet}. Counterfactual explanations

\begin{table*}[hbt!]
\caption{Taxonomy of visual explanations for end-to-end autonomous driving. Classification is partially based on  \cite{zablocki2022explainability}.}
\label{tab:visual_explanations}
\setlength\tabcolsep{5pt}
\setcellgapes{3pt}
\makegapedcells
\begin{tabular}{
    >{$}p{\dimexpr0.20\linewidth-2\tabcolsep-1.20\arrayrulewidth}<{$}
    p{\dimexpr0.40\linewidth-2\tabcolsep-1.46\arrayrulewidth}
    >{\centering\arraybackslash}p{\dimexpr0.40\linewidth-2\tabcolsep-1.33\arrayrulewidth}
}
\hline
\rowcolor{lightgray}
\multicolumn{1}{>{\centering\arraybackslash}p{\dimexpr0.20\linewidth-2\tabcolsep-1.20\arrayrulewidth}}{\textbf{\textcolor{Cyan}{Type of explanation}}} &
\multicolumn{1}{>{\centering\arraybackslash}p{\dimexpr0.39\linewidth-2\tabcolsep-1.46\arrayrulewidth}}{\textbf{\textcolor{Cyan}{Description of explanation}}} &
\multicolumn{1}{>{\centering\arraybackslash}p{\dimexpr0.38\linewidth-2\tabcolsep-1.33\arrayrulewidth}}{\textbf{\textcolor{Cyan}{Relevant studies}}} \\
\hline
\multicolumn{3}{c}{\textit{Post-hoc explanations}} \\
\hline
\centerline{{Saliency maps}} & {Indicate what part of an input image has more influence in the prediction of the model. This approach primarily includes back-propagation, local approximation, and perturbation-based methods.} & {\cite{bojarski2016end, bojarski2017explaining, bojarski2018visualbackprop, mohseni2019predicting, kim2017interpretable, kim2021toward, sauer2018conditional, liu2020interpretable}} \\
\hline
\centerline{{Counterfactual explanations}} & {Indicate explanatory information on changing input slightly and observing the causal effect of this modification on the prediction of the model.} & {\cite{bansal2019chauffeurnet, li2020make, jacob2022steex, zemni2023octet}} \\
\hline
\centerline{{Model translations}} & {Transfer knowledge from a black-box neural network entirely to a more interpretable model.} & {Not available} \\
\hline
\centerline{{Representation explanations}} & {Explain intermediate representations, i.e., the internal structure of the model.} & {\cite{tian2018deeptest}} \\
\hline
\multicolumn{3}{c}{\textit{Intrinsic explanations}} \\
\hline
\centerline{{Built-in attention models}} & {Provide reasoning on the inner workings of the learning model via built-in attention maps.} & {\cite{kim2017interpretable, lee2017desire, wang2019deep, kim2020attentional, araluce2024leveraging}} \\
\hline
\centerline{{Semantic inputs}} & { Interpretable input spaces with dimensions individually interpretable and concrete meaning.} & {\cite{bansal2019chauffeurnet, cui2019multimodal, djuric2020uncertainty}} \\
\hline
\centerline{{Auxiliary information}} & {Additional information, different for driving model its, that provides human-interpretable information during the trip (e.g., a semantic map of road objects and users in a scene).} & {\cite{mehta2018learning, zeng2019end, sadat2020perceive}}
\\
\hline
\centerline{{Representation visualization}} & {Describes information contained inside the intermediate representation.} & {\cite{morton2017simultaneous, hu2022st}}
\\
\hline
\centerline{{Natural language explanations}} & {Human interpretable linguistic explanations on the predictions of the model.} & {\cite{kim2018textual, omeiza2021towards, ben2022driving, 10421901, feng2023nle}}
\\
\hline
\end{tabular}
\end{table*}\hspace{-0.53cm}
are in general well-aligned with causality and causal inference \cite{pearl2009causality}. Another potential gain
from a safety perspective of counterfactual explanations is that foreseeing potential outcomes of different choices provides a chance to learn from mistakes, avoid high-stakes decisions, and make better choices in other similar traffic circumstances. \\
\textit{(1.2) {Global Explanations:}} Global explanations refer to explanations that provide a summary of the decision-making process of an entire learning model \cite{longo2020explainable}. 
In context of end-to-driving, global explanations remain scarce.  Tian et al.'s work \cite{tian2018deeptest}, to the best of our knowledge, is the sole example provide an automated testing of end-to-end driving. They use the idea of \textit{neuron coverage} to test deep neural networks-based  autonomous driving models, named \textit{DeepTest} as the neuron coverage concept is regarded as a logic quantification technique that is exploited by a set of testing inputs.
\subsubsection{Intrinsic Explanations}
Finally, except for explaining the prediction of a model in a post-hoc manner, there are attempts to develop models that are \textit{interpretable by design}. In the context of autonomous driving, these explanation-by-design systems include a variety of ideas, including inherent attention models, semantic inputs, auxiliary information, representation visualization, and live natural language explanations.  Table \ref{tab:visual_explanations} describes the relevant studies on post-hoc and intrinsic explanations for end-to-end driving. 
\subsection{Reinforcement learning-based Explanations }
Explainable reinforcement learning (XRL) is a considerably new, emergent research area \cite{heuillet2021explainability,vouros2022explainable} and has recently been applied to autonomous driving tasks. XRL approaches for

\hspace{-0.45cm}
end-to-end driving can be categorized via feature importance \cite{wang2021interpretable, hejase2022dynamic, shao2023safety, li2025explainable, cui2022interpretation},  policy-level explanations \cite{guo2021edge}, \cite{kenny2023towards}, reward-decomposition \cite{yang2023leveraging}, logic-based explanations \cite{song2022interpretable}, and saliency maps {\cite{shi2020self, chen2021interpretable, bellotti2022designing}}. It is noteworthy to specify that RL-based explanations are primarily targeted at researchers and developers (rather than public users, such as passengers) for intrinsic model interpretability. Such explanations, as log data, can provide insights into why specific actions are taken, thus enabling pinpointing potential issues in the RL model and refining it for better performance. (see Table \ref{tab:RL_explanations}). We present insights from an RL-based explainability perspective in Experiment 2, Section VI.B.

\begin{table*}[hbt!]
\caption{Taxonomy of explainable reinforcement learning approaches for end-to-end autonomous driving}
\label{tab:RL_explanations}
\setlength\tabcolsep{5pt}
\setcellgapes{3pt}
\makegapedcells
\begin{tabular}{
    >{$}p{\dimexpr0.30\linewidth-2\tabcolsep-1.20\arrayrulewidth}<{$}
    p{\dimexpr0.38\linewidth-2\tabcolsep-1.46\arrayrulewidth}
    >{\centering\arraybackslash}p{\dimexpr0.32\linewidth-2\tabcolsep-1.33\arrayrulewidth}}
\hline
\rowcolor{lightgray}
\thead{\textbf{\textcolor{cyan}{Type of explanation}}}  
& \thead{\textbf{\textcolor{cyan} {Description of explanation}}} 
& \thead{\textbf{\textcolor{cyan} {Relevant studies}}} \\  
\hline
\centerline{{Feature importance}} 
& Describes features/states enabling an RL agent to take a specific action.
& {\cite{wang2021interpretable, hejase2022dynamic, shao2023safety}} \\
\hline
\centerline{{Policy-level explanations}} 
& Policies that are interpretable by design.
& {\cite{schmidt2021can,paleja2022learning, tambwekar2023natural, kenny2023towards}} \\
\hline
\centerline{{Reward-grounded explanations}} 
& Provide reward-based insights on an RL agent's actions. 
& {\cite{yang2023leveraging}} \\
\hline
\centerline{{Logic-based explanations}} 
& Neuro-symbolic and neural logic models to learn interpretable controllers. 
& {\cite{song2022interpretable}} \\
\hline
\centerline{{Saliency maps}} 
& Visualization of an RL agent's actions for behavior understanding. 
& {\cite{shi2020self, chen2021interpretable, bellotti2022designing}} \\
\hline 
\end{tabular}
\end{table*}
\hspace{-0.48cm}
\subsection{Imitation learning-based Explanations}
Similar to RL-based explanations, IL-based explanations also provide insights from an algorithmic perspective regarding how an IL agent makes decisions. Cultrera et al. 's work \cite{cultrera2020explaining} is one of the preliminary studies in this context. The study uses a conditional IL agent along with the attention model in the CARLA simulator \cite{dosovitskiy2017carla} to produce an explanation describing the influence of distinctive parts of the driving images on the predictions of the model. Further works include NEAT \cite{chitta2021neat}, model-based IL (MILE) \cite{hu2022model},  Hierarchical Interpretable Imitation Learning (HIIL) \cite{teng2022hierarchical}, PlanT \cite{renz2023plant}, and generative adversarial network (GAN)-based IL approach combined with signal temporal logic (STL) \cite{liu2024interpretable}.

\subsection{Feature Importance-based Explanations}
Quantitative metrics provide a feature importance-based interpretability approach and have recently been explored in
autonomous driving. These studies primarily use decision tree-based metrics and SHAP values \cite{lundberg2017unified} to quantify the contribution of each feature in predictions of a driving model, thus providing a holistic understanding of how inputs influence an AV's real-time actions. Examples include vehicle positioning \cite{onyekpe2022explainable, almalioglu2022deep}, risk assessment \cite{nahata2021assessing}, driver takeover time \cite{ayoub2022predicting}, anomaly detection \cite{nazat2024xai}, trust management \cite{rjoub2022explainable} with SHAP, and intelligible explanation provision with tree-based representation \cite{omeiza2021towards}.
\subsection{Pretrained Large Language Models and Vision-Language Models-based Explanations}

Pretrained large language models (LLMs) and vision-and-language models (VLMs) have recently revolutionized natural language processing (NLP), robotics, and computer vision across many tasks. Primarily based on BERT \cite{devlin2019bert} and GPT \cite{radford2018improving} families, there has been a surge in building general-purpose language models. These models are called \textit{Foundation Models} where their internal weights are further adjusted (i.e., ``fine-tuned'') with a domain-specific knowledge base for serving task-specific purposes. LLMs and VLMs in explainable autonomous driving have recently been investigated from several aspects. For example, the Talk2BEV \cite{dewangan2023talk2bev} model has incorporated a VLM into bird’s-eye view maps, enabling spatial and visual analysis to predict unsafe traffic scenarios. In a similar work, DriveGPT4 \cite{xu2023drivegpt4} justifies AV decisions via textual descriptions and responds to humans’ questions in the same manner. The concept of question-answering as a human-machine interaction has been explored in two further models LingoQA \cite{marcu2023lingoqa} and VLAAD \cite{park2024vlaad}.  Table \ref{tab:llvm} presents recent LLM and VLM-based explainability approaches for end-to-end driving. It is also noteworthy to underscore that the current well-known limitations of these large models (i.e., ``hallucinations,'' authoritative but incorrect explanations) can have dire consequences for understanding driving safety in real driving environments. Section VI sheds light on these caveats with empirical analyses and describes the implications of LLM- and VLM-generated explanations for safety of end-to-end driving.  

\subsection{Time Granularity and Faithfulness of Explanations}
The time sensitivity of explanations is yet another essential factor from a safety perspective. Explanations, depending on
the level and context they are delivered, can have different time granularity, ranging from milliseconds to seconds and sometimes to a longer interval. In general, the time granularity of explanations in AVs could be grouped as follows:\\
\textit{Immediate feedback (in milliseconds)}: Some explanations may have higher time sensitivity and need to be delivered in the range of milliseconds. For instance, if an AV encounters a sudden obstacle or an unexpected object at a very near distance, it should provide immediate feedback to the human passengers or backup driver for a possible takeover or situation awareness. This feedback accompanied with a relevant explanation could describe why the vehicle made a specific maneuver or applied an emergency brake, for example. \\
\textit{Perceptual explanation (instantly and in a few seconds)}: Explanations for an AV’s sensing of the operational environment might have been provided instantly and within a slightly longer time allowance, typically in the range of seconds, depending on the time-criticality of the situation. For instance, saliency-based visual explanations could highlight regions in the image that contributed most to the AV’s instant action decision. On the other hand, some scenarios could have a higher degree of tolerance in terms of time for perceptual explanations (i.e., detection of an unexpected static obstacle in a drivable area within around a hundred meters). Other examples may be the justifications for these vehicles' detection of a pedestrian, a stop sign ahead, or another vehicle in their vicinity. Perceptual explanations within a range of seconds could be helpful for

\begin{figure}
\includegraphics[width = \columnwidth]{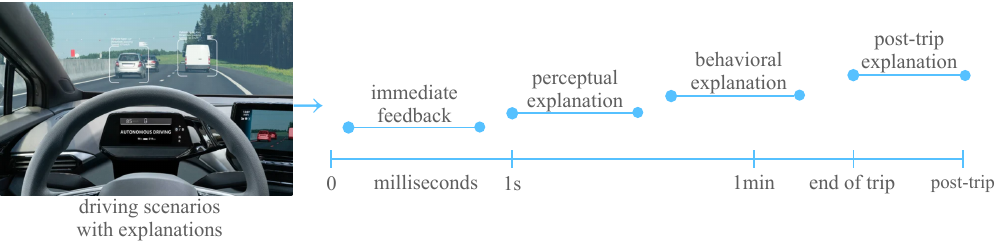}
\caption{The time length of communicating explanations for safe and on-time human, vehicle reactions and situation awareness in autonomous driving.}
 \label{fig:time_granularity}
 \end{figure}

\begin{table*}
  \centering
  \caption{Pretrained large language and vision models for explainable autonomous driving}
  \label{tab:llvm}
  \begin{tabularx}{\linewidth}{>{\raggedright\arraybackslash\hsize=0.30\hsize}X >{\raggedright\arraybackslash\hsize=0.70\hsize}X}
    \toprule
    \multicolumn{1}{l}{\textbf{\hspace{0.5cm}\textcolor{cyan}{Pretrained language and vision model}}} & \multicolumn{1}{l}{\textbf{\hspace{3cm}\textcolor{cyan}{Description of the model}}} \\
    \midrule
    \hspace{1.5cm}Talk2BEV \cite{dewangan2023talk2bev} & Predicting behavior of traffic actors and events via  visual cue and reasoning \\
    \addlinespace
    \hspace{1.5cm}DriveGPT4 \cite{xu2023drivegpt4} & Providing textual explanations for AV actions and question-answering interface for humans \\
    \addlinespace
    \hspace{1.5cm}LingoQA \cite{marcu2023lingoqa} & Video Question Answering for traffic scene understanding \\
    \addlinespace
    \hspace{1.5cm}Driving with LLMs \cite{chen2023driving} &Context understanding in traffic scenarios with vectorized numeric modalities \\
    \addlinespace
    \hspace{1.5cm}GPT-Driver \cite{mao2023gpt} & Interpretable motion planning with a language model\\
    \addlinespace
    \hspace{1.5cm}DILU \cite{wen2023dilu} & Commonsense knowledge-based decision-making via Reasoning and Reflection modules \\
    \addlinespace
    \hspace{1.5cm}LanguageMPC \cite{sha2023languagempc} & Chain-of-thought-based reasoning for decision-making and control \\
    \addlinespace
    \hspace{1.5cm}LINGO-1 \cite{wayve_team} & Providing live textual explanations for AV actions \\
    \addlinespace
    \hspace{1.5cm}VLAAD \cite{park2024vlaad} & Traffic scene understanding from a frontal view of an AV using video question answering \\
    \addlinespace
    \hspace{1.5cm}DriveLM \cite{sima2025drivelm} & Proxy task to mimic the human reasoning process and enabling interactivity with human users \\
    \addlinespace
    \bottomrule
  \end{tabularx}
\end{table*}\hspace{-0.48cm}
takeover requests such as the blind corner case described in Figure \ref{fig:take_over}. In the experiments for a lead time between 4 seconds vs. 7 seconds, Huang and Pitts \cite{huang2022takeover} have shown that a shorter lead time leads to faster reaction but the poor quality of takeover.  On the other hand, people’s age could be another critical factor, and older people may need more reaction time in takeover situations \cite{wiseman2024autonomous}. However, when it comes to an AV, this time is roughly 100 milliseconds \cite{xu2012real}. While the reaction time of 100 milliseconds for an AV may seem impressively fast, it can still be insufficient in complex driving situations due to possible communication latency and physical constraints of the vehicle. The optimal lead time in takeover is indeed scenario-dependent and may involve a trade-off from the effectiveness-stress perspective. Ideally, longer times may give humans enough time to effectively manage the takeover situation; however, when this time is too short while requiring instant intervention, it can cause stress for a human actor, as a downside. Overall, perceptual explanations with appropriate timing can bring situation awareness for people on board. \\ 
\textit{Behavioral Explanation (seconds to minutes):} These explanations may reflect the rationale behind route planning or a longer-term decision-making process. Behavioral explanations may be provided in advance to inform passengers of upcoming changes in the journey and help them understand the vehicle's decision-making intentions. \\
\textit{Post-trip explanations (minutes, days, weeks, years):} Once the AV completes its journey, the explanation system can provide a detailed post-trip analysis, explaining the decision-making processes from the initial point to the final destination, highlighting critical events along the trip, and offer insights into how the system performed and what can be improved. In general, the choice of time granularity for explanations should take the urgency, relevance, and potential impact of the information conveyed into account. Real-time safety-critical information requires immediate explanations, while lower-priority longer-term decisions and post-trip explanations can be communicated with a longer time horizon. Figure \ref{fig:time_granularity} shows the timing sensitivity of explanations in this regard. \\
Regardless of whether explanations are based on informational content or timing perspective, faithfulness of explanations
is
of paramount importance. Consequently, explanations in autonomous driving should be evaluated carefully and such evaluation should also present insights from their safety perspective. In this sense, Kim et al. \cite{kim2023and} have investigated the role of on-road explanations and evaluation approaches for them from timing and delivery-type perspectives in highly autonomous driving. They provide three types of explanations, namely \textit{perception}, \textit{attention}, and \textit{perception+attention} in a windshield display of a vehicle, and evaluate explanations both in a laboratory setting and on actual roads. The empirical findings of the study suggest risk-adaptive explanations were confirmed to be more effective when driving-related information is overwhelming to passengers on board and such explanations also enhance passengers’ overall trust and perceived safety in an AV. It turns out that the ``what” and ``why” facets of explanations provide more value in critical scenarios for passengers as safety implications in a highly autonomous driving environment.
 
\section{Insights from Analytical Case Studies}

Having covered safety principles and  XAI approaches with critical explanation aspects for end-to-end driving, we combine these concepts for safety analysis hereafter, which is the main focus of our paper. In this sense, this section presents analytical studies showing how XAI can improve the safety of end-to-end driving both in real-time and via a retrospective analysis. 

\subsection{Real-time Explanations for Safety Monitoring} 

Explanations provision can be in real time or in a post-hoc manner, depending on the task and application domain. In autonomous driving, as decisions are temporal and safety-critical, real-time explanations of the AV decisions can help the users (i.e., backup drivers and passengers) monitor driving

\begin{figure*}[htp]
    \centering
    \includegraphics[width=18cm]{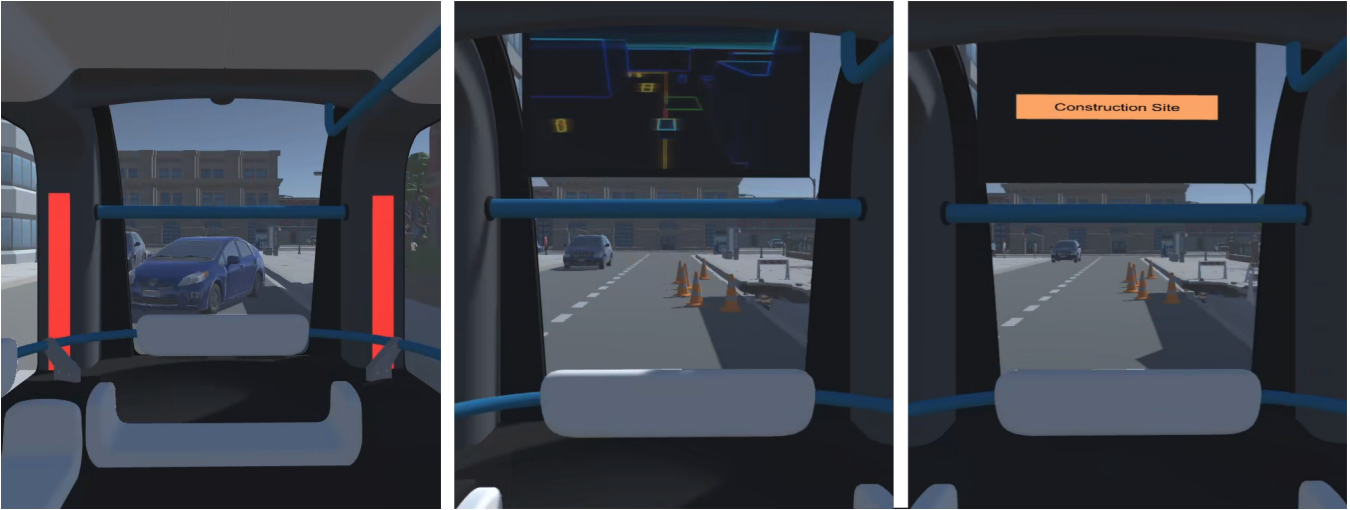}
    \caption{Temporal evolution of three driving scenes and information conveyance to passengers using visual and textual explanations at these scenes, based on the AVs' closeness to the other vehicles in oncoming traffic. Graphics credit: Schneider et al., \cite{schneider2021increasing}.}
    \label{fig:three_scenarios}
\end{figure*}\hspace{-0.44cm}
safety and possibly intervene in the situation in case the driving system malfunctions. In this context, an AV must be equipped with the relevant user interface or dashboard that conveys scene-based information to in-vehicle users, and possibly an emergency control button, which passengers can use in case of unsafe actions. Such a user interface has two main safety implications for end-users. First, information displayed in the user interface can communicate situation awareness and help users trust the vehicle. Furthermore, explanations can also help understand the intentions of a vehicle in its decisions. An interesting study in this perspective has been carried out by Schneider et al. \cite{schneider2021increasing}. The authors use multimodal explanation feedback using light, sound, visual, vibration, and text formats to observe whether such techniques make a positive user experience in their human study. They categorize the design of explanations
with the help of autonomous driving domain experts and researchers. Based on this approach, it is concluded that driving scenario-based feedback can be four types depending on how critical, reactive, and/or proactive the scenario is, and three of them are illustrated in Figure \ref{fig:three_scenarios}:\\
1. \textit{Proactive non-critical scenarios}: An AV does not perform a dangerous action and has enough time to react to a scene. The situations in the middle and rightmost segments of Figure \ref{fig:three_scenarios} are representatives of proactive non-critical scenarios.  \\
2. \textit{Proactive critical scenarios}: This is when the situation is sufficiently hazardous and a vehicle is expected to act on time to avoid any potential danger or a mishap. \\
3. \textit{Reactive non-critical scenarios}: The vehicle should act immediately to avoid a situation that does not imperil human lives. An example is the situation where an animal runs across the road from an invisible place in front of the vehicle. In this case, the vehicle must press the brake and prevent hitting. \\
4. \textit{Reactive critical scenarios}: There is insufficient reaction time that has a high degree of risk or danger for human life. An example is a situation in which another vehicle appears in oncoming traffic at a close distance. The vehicle should immediately brake to avoid a potential collision. The leftmost image segment in Figure \ref{fig:three_scenarios} is a relevant example. \\
These four categories of driving scenes show that real-time explanatory information on driving scenes has significant safety implications for autonomous driving. Explanations provide driving scenario-based information to the passengers and show how the vehicle behaves in such circumstances. Moreover, they can also communicate the vehicle's existing limitations, issues, and enable debugging and enhancing the driving system.  

\subsection{Failure Detection with Explanations}
\begin{figure*}[htp]
    \centering
    \includegraphics[width=1\textwidth]{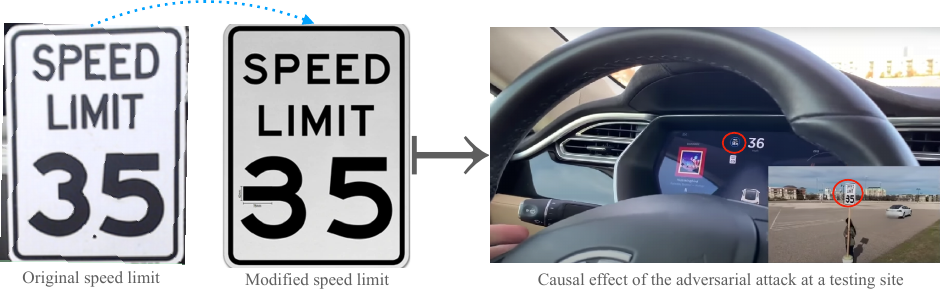}
    \caption{A deliberate hack causes the 35-mph sign limit to be incorrectly perceived as an 85-mph sign by Tesla's ADAS at a testing site. The manually added red circles show the speed limit perceived by the heads-up display and modified speed sign. The figure has been drawn based on the content in \cite{povolny2020model}.}
    \label{fig:tesla_hack}
\end{figure*}

An ability to explain incorrect actions or autonomous driving failures is another positive and significant aspect of vehicle autonomy. The reasons for failures could originate from individual or multiple sources, such as software bugs, sensor malfunctions, communication breakdown, poor roads, bad weather conditions, and adversarial machine learning attacks as specified in \cite{cui2019review}. Foreseeing such potential safety threats is also crucial in the design, development, and continual debugging of self-driving systems. 
In particular, testing AVs in their prototype stage can provide ample opportunities to understand the functionalities and limitations of the driving system and certify compliance with the V-model of the ISO 26262 standard. In this context, McAfee researchers have conducted an experiment, termed \textit{model hacking}, to test the vision system of Tesla Model vehicles against adversarial attacks \cite{povolny2020model}. They made some alterations to some traffic objects, such as the speed limit, to observe how the perception system of an AV understands the modified image and acts in the driving scene. For this purpose, the team deliberately modifies the speed limit as shown in Figure \ref{fig:tesla_hack}: They added a black sticker to the middle of 3 in the speed limit and tested Tesla Model S at a site to see how the heads-up display of the vehicle reads the altered speed limit. As shown, the vehicle reads the 35-mph sign as an 85-mph sign and accelerates once it approaches and passes the sign. Such misdetection, particularly given that the difference between the modified and original speed limits is substantially big, could have dire consequences in a real driving environment.\\ 
With that said, even without a more careful look, an ordinary person may not understand why the AV accelerated in this case, as the alteration in the speed limit is quite deliberate and not easily visible. In this case, a more appropriate approach to interpreting the vehicle’s decision is to acquire a causal explanation. In such scenarios, both the real-time and post-hoc explanations can bring safety benefits as shown below:\\
\textit{Real-time explanation}:  If the AV provides a rationale for its decision on a dashboard or user interface (such as “The speed limit is 85 mph, accelerating”) in real-time while approaching the speed sign, a passenger on board or a backup driver may understand such a decision and possibly intervene in the situation by pressing the emergency stop button and decreasing the speed. Timely communication of the malfunctions of the vehicle to the users may help avoid potential dangers ahead.\\
\textit{Post-hoc explanation}: In case the explanations are delivered in a post-hoc manner, a history of action-explanation pairs could help detect correct/incorrect actions through debugging.

\subsection{Solving the "Molly Problem" with Explanations}
\begin{figure}[t!]
\includegraphics[width = \columnwidth]{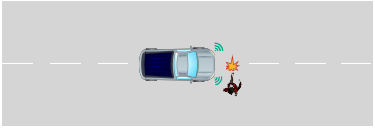}
\caption{The Molly problem: An AV hits a pedestrian and nobody witnesses. Explainability of the AV decisions can help understand why the AV kept going and eventually hit the pedestrian at this scene.}
 \label{fig:molly}
 \end{figure}

As AVs are becoming highly dependent on their automotive features, several safety challenges evolve with their black-box AI-based decisions. Road accidents with such vehicles can trigger a variety of regulatory inspections from safety, engineering, ethical, and liability perspectives. Notably, one of the most debatable issues within such a context is the proper investigation of AV-involved collisions and hitting cases where there are no eyewitnesses. In this sense, ADA Innovation Lab Limited and the Technical  University of Munich researchers have formalized the “Molly problem” (i.e., Figure \ref{fig:molly}), which addresses the ethical challenges when an unoccupied AV hits a person with no witness in the scene \cite{molly2020}. The Molly problem describes a traffic scenario with an event of hitting a person and seeks an answer to the following question: \\

\begin{quote}
 \textit{A young girl called Molly is crossing the road alone and is hit by an unoccupied self-driving vehicle. There are no eye witnesses. What should happen next?}  
\end{quote}
\vspace{0.2cm}
Such a situation is considerably challenging, and the primary goal of the post-accident inspection is to identify which part made a mistake: Was that a faulty decision made by the self-driving vehicle, or did Molly unexpectedly enter the driving zone and cause the mishap? To cope with the issue, the research team has created a survey to get public views on this problem to identify the main culprit of the road accident. 296 respondents aged between 18 and 73 years old were asked questions about the driving software of the AV and its impact on hitting \cite{molly_survey}. 75\% of these people reported that they favored traveling in AVs in general. According to the survey result, 97\% expected that the AI software of the AV should be aware of the hitting, and 94\% of the respondents believed that the software should have stopped the AV at the collision area. Moreover, 94\% thought that the AV should have indicated a hazard signal to bystanders on the scene. It turns out that a majority of societal views hold AVs more responsible for such accidents (Table \ref{table_molly_survey}). \\
Such traffic accidents with implications from safety to liability necessitate the concept of explainability of the driving system to a substantial degree. Given that nobody witnessed the collision, it seems that only an accurately delivered history of action-explanation log data could be helpful for forensic analysis and understanding of the main cause of the mishap.  Furthermore, as the collision process evolves over a short period of time, we propose that explainability of the driving system may be analyzed over three time phases with the following questions: \\
\textit{Phase 1 - Before hitting:} Did the AV follow
the traffic rules (such as the speed limit) on that road segment and detect the pedestrian before the collision? If so, just before the collision, did the vehicle press the emergency brake even though the hitting eventually became inevitable? \\
\textit{Phase 2 - At the hitting point:} Once the accident occurred, did the vehicle “understand” that it hit a person and come to a full stop accordingly, as an expected course of action?\\
\textit{Phase 3 - After hitting:} If the vehicle became aware of hitting, did it activate emergency state functions such as reporting the accident to the regulatory bodies and emergency service immediately? 

\begin{figure*}[htp]
    \centering
    \includegraphics[width=1\textwidth]{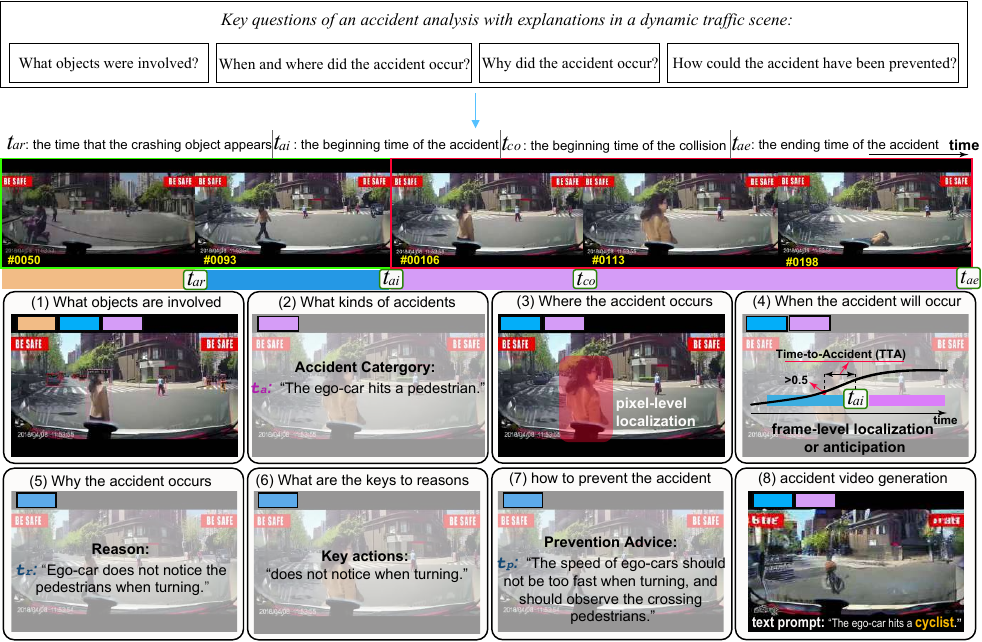}
    \caption{An example of object-centric accident video understanding from the Multi-Modal Accident Video Understanding (MM-AU) dataset of Fang et al. \cite{fang2024abductive}. The graphics describe the temporal evolution, the root cause of the accident, and accident prevention advice in different time steps with textual explanations.}
    \label{fig:accident_analysis_by_Fang}
\end{figure*}
 \hspace{-0.48cm}
Getting answers to these is vital for a proper post-accident inspection. For instance, referring to Phase 2, the case where the AV stopped once the hitting happened would have had different legal implications than the not-stopping scenario, which can help deal with the arising ethical, responsibility, liability, and accountability issues accordingly.
\subsection{Spatiotemporal Object-Centric Accident Analysis with Explanations}
The Molly problem mentioned above is a typical scenario where the role of explanations becomes indispensable. The immense need for more fine-grained explanations becomes even more vital in more complex scenarios involving several dynamic and stationary objects \cite{fang2024abductive, zhang2021exploring}. Thus, to prevent

\begin{table}[htb]
\caption{The Molly problem survey: Participants' answers to the selected safety-related queries. The table reproduced based on \cite{molly_survey}.}
\begin{center}
    \begin{tabular}{c c c c}
\hline
{\cellcolor{white}{\textcolor{cyan}{Query}}} & {{\textcolor{cyan}{Expected}}}& \textit{\textcolor{cyan}{Unsure}}& \textit{\textcolor{cyan}{Didn't expect}} \\
\hline
\makecell{\scriptsize{AI software should be} \\ \scriptsize{aware of the collision.}} & 97\%&2\% &1\%  \\
\hline
\makecell{\scriptsize{AI software should stop} \\ \scriptsize{the AV at the collision area.}} & 94\%&4\% &2\%  \\
\hline
\makecell{\scriptsize{AI software should indicate} \\ \scriptsize{a hazard to road users.}} & 97\%&2\% &1\%  \\
\hline
\end{tabular}
\label{table_molly_survey}
\end{center}
\end{table} \hspace{-0.45cm}
traffic accidents, relevant explanations should clearly describe what objects are involved, how the accident evolved over time, and how it can be prevented. In this sense, we refer to Fang et al.’s recent study \cite{fang2024abductive} on spatiotemporal analysis of accidents involving several traffic actors. Figure \ref{fig:accident_analysis_by_Fang}, showing a traffic scene from their MM-AU dataset \cite{fang2024abductive}, describes the objects involved, the type of accident, the place and timing of the accident, the key reason for the accident, and how to prevent such a mishap as a general object-centric accident analysis and prevention method. The example clearly outlines the temporal evolution of the accident with causal reasoning. The accident can be analyzed differently from the modular and end-to-end driving perspective. While the root cause of the accident can easily be linked to a \textit{ specific module} (e.g., "The perception module failed to detect the pedestrian due to overexposure to the sunlight ahead") for modular driving, in the end-to-end setting, explanations can reveal \textit{latent patterns} in training data that led to failures (e.g., the model ignored pedestrians in certain lighting conditions), identify gaps in training datasets, and thus reveal \textit{data-driven insights} about the accident.
\subsection{Overall Analysis of the Case Studies Above from the Driving Paradigm Perspective: End-to-end vs. Modular Pipeline }
The analysis performed in the above section also applies to the other three case studies and helps draw a unique conclusion to differentiate end-to-end driving from modular driving regarding safety impacts of the explanations: While explanations can reveal faulty behavior by attributing it to a specific \textit{module} in modular driving with an easy way to diagnose failures (as each module has defined responsibilities), the main implication of explanation would be a \textit{data} perspective for end-to-end driving: Targeted data augmentation could help refine the end-to-end model’s behavior and improve safety performance of the driving system, given that end-to-end driving is more data-hungry, and supplying more relevant ``less-seen" data for the revised model can help end-to-end driving be safer in interacting with dynamic environments within embodied AI.
\section{Insights from Experimental Analyses}
Having described analytical studies on the safety analysis of explanations in Section V, this section delineates insights from quantitative and qualitative empirical studies and shows implications of the faithfulness of explanations from a safety perspective practically as a follow-up. It presents insights from two experimental analyses - qualitatively and quantitatively - and describes critical findings from these experiments.

\subsection{Experiment 1: Qualitative Analysis on Video Question Answering as Action Explainer}
\begin{figure*}[htp]
    \centering
    \includegraphics[width=1\textwidth]{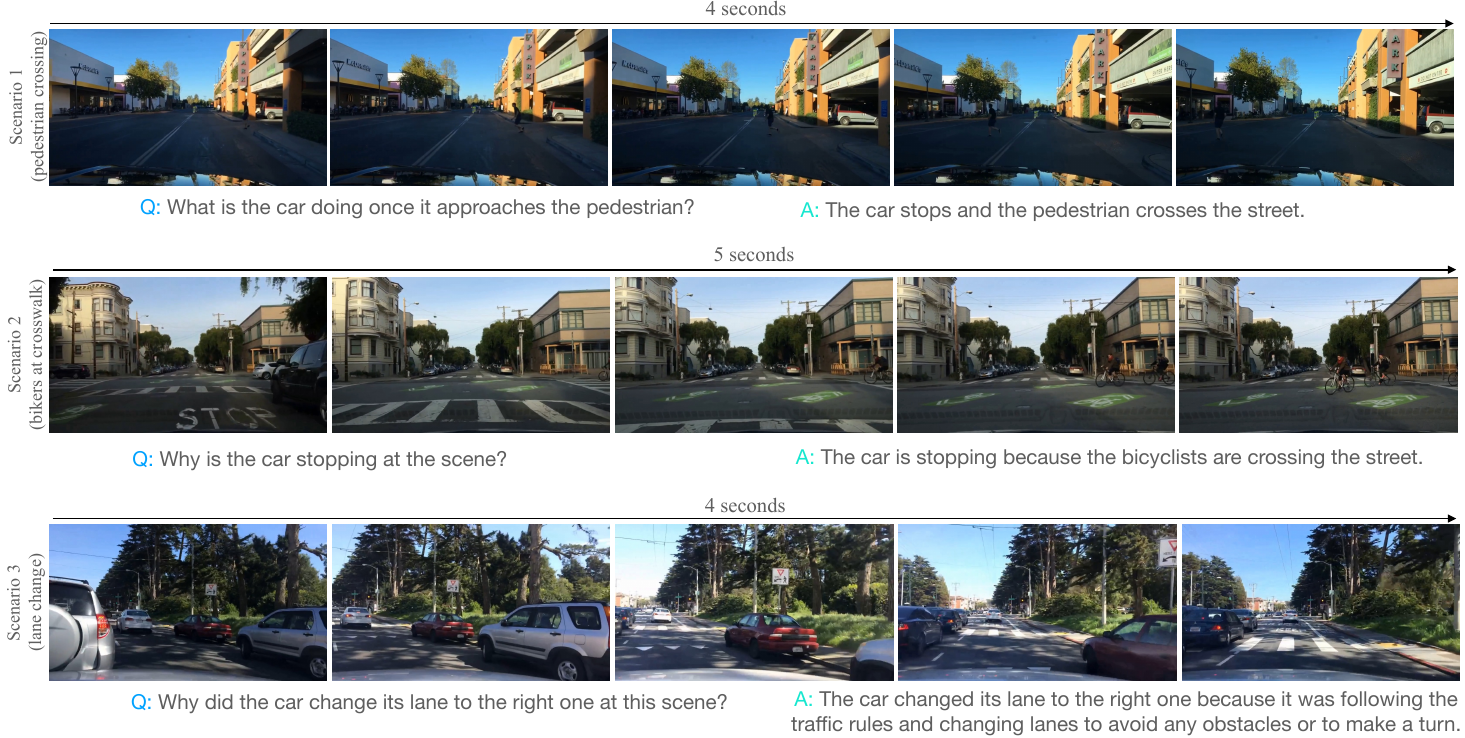}
    \caption{The results of our experiment on the chosen scenarios from the BDD-A dataset (recorded videos) with the Video-LLaVA multimodal transformer, as an explanation model. The model produces correct explanations for the conventional questions on the AV actions in the described scenes.}
    \label{fig:BDDA}
\end{figure*}

\begin{figure*}[htp]
    \centering
    \includegraphics[width=1\textwidth]{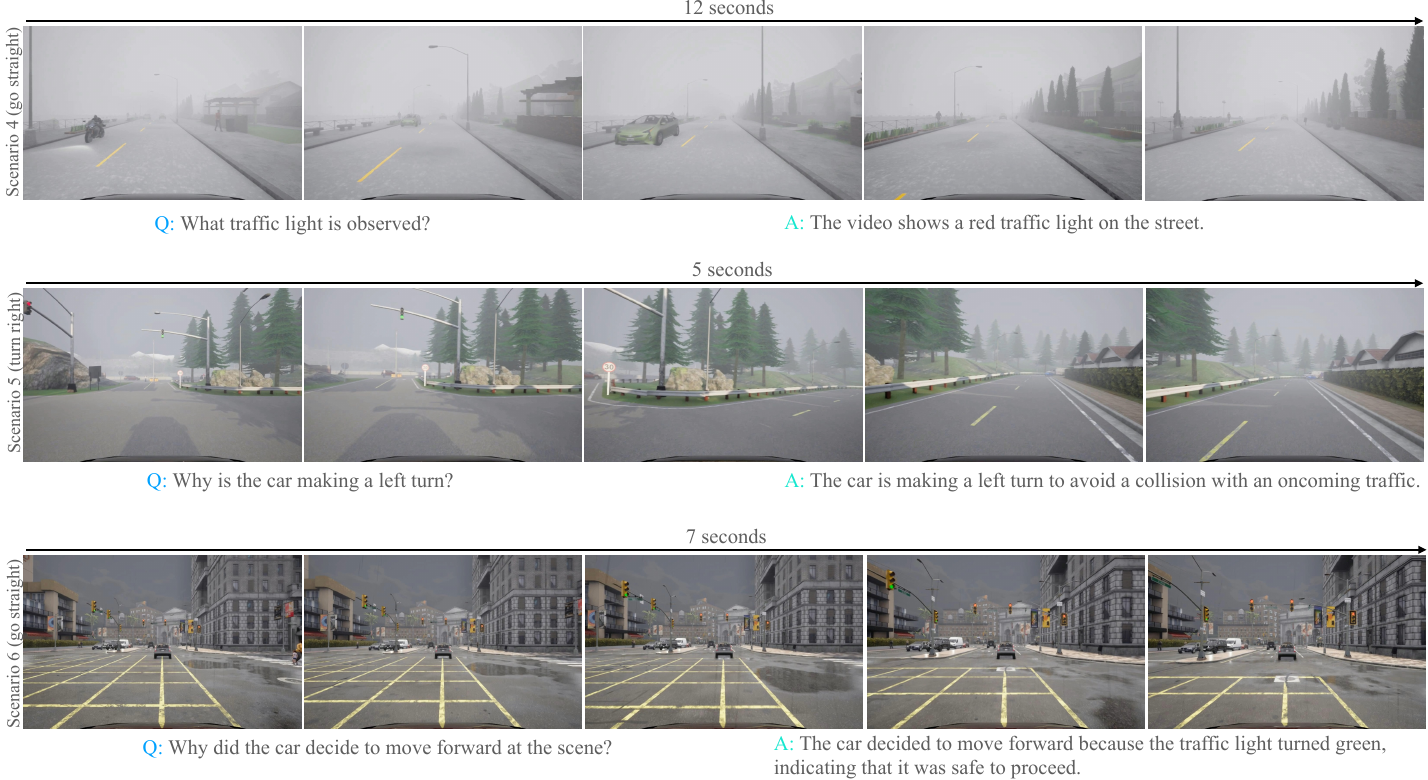}
    \caption{The results of our experiment on the chosen scenarios from the SHIFT dataset (recorded videos) with the Video-LLaVA multimodal transformer, as an explanation model. Our deliberate questions confuse the model: In Scenarios 4 and 5, the model is influenced by tricky questions and generates incorrect responses. In Scenario 6, the explanation model fails to provide an adequate response on why the AV kept going straight under the red light.}
    \label{fig:SHIFT}
\end{figure*}

This subsection describes our experiment with video question answering (VideoQA) applied to real-driving and simulation datasets, as an explanation provision method in end-to-end driving. In this sense, we extend our recent work \cite{10421901} on explaining autonomous driving actions via a visual question answering (VQA) mechanism. While our preliminary work focuses on providing textual explanations for a visual scene in a stationary environment with a single driving image, we expand the scope of that work, evaluate explanations in dynamic environments with video-based driving scenes, and describe implications of such explanations from a safety point of view. The key idea with the video question answering is that an explanation must capture the semantics of the temporal changes in visual driving scenes. We sample six driving scenes from recorded driving videos of the SHIFT,  a simulation
dataset \cite{sun2022shift}, and Berkeley DeepDrive Attention (BDD-A), a real-world driving dataset \cite{ xia2019predicting}. The lengths of these scenarios vary from 4 seconds to 12 seconds. We use the Video-LLaVA \cite{lin2023video} multimodal transformer as an explanation mechanism for our VideoQA task, which takes a driving video and a question about the context of this video as input and produces a response.  While video-based multimodal LLM for explaining AV actions could simply be a \textit{plug-and-play}
or \textit{driving-dedicated} model, in this experiment, we consider Video-LLaVA as a plug-and-play video-reasoning model (e.g., the AV might have a different mental model than the video reasoning model). So, our explicit focus with this experiment is directly on the implications of explanations generated by the video reasoner, regardless of whether such a reasoner has been pretrained on driving or general domain data. Such a task, in general, can be considered as an interactive dialogue between humans and a conversational user interface where people on board ask questions to the user interface, and LLM-based temporal explanations can help with situation awareness in dynamic traffic scenes and thus improve the understanding of AV's actions. \\
In our experiment, we have carefully selected scenarios and asked purposeful questions to see how an explanation model responds to our question. Overall, we can analyze safety implications of action-explanation pairs in four ways: \\
\textit{1. Correct action, correct explanation}: This case is a desideratum as the ultimate goal of explainable autonomous driving is to choose and perform actions correctly and provide action-reflecting and context-aware explanations as required by the safety-regulatory compliance principles. Scenarios 1, 2, and 3 in Figure \ref{fig:BDDA} may be deemed as examples of such a category. \\
\textit{2. Correct action, incorrect explanation}: The question-answer pair in Scenario 5 falls into this category. When we ask an \textit{adversarial} question as “Why is the vehicle making a left turn?", the model is flawed and generates a falsified response of “The vehicle is making a left turn to avoid a collision with an oncoming traffic,” where the AV, in fact, performs a \textit{right turn}. During the trip, users may ask \textit{incorrect} questions either unintentionally (i.e., visually impaired passengers) or intentionally/deliberately (to test robustness of the explanation system). Accordingly, the explanation interface must detect such questions and respond correctly. Failing to provide adequate responses to adversarial questions may damage the users' trust in the explanations and even action decisions of a self-driving vehicle. A rigorous VideoQA model must not only provide action/scene-related explanations but also defend against adversarial questions to ensure that the vehicular interface delivers an explanation of what it perceives in the scene. Moreover, robustness against human adversarial questions can contribute to the development of safety and security measures to detect and mitigate potential adversarial attacks on the perception system of an AV. \\
Another potential limitation of an explanation model could be its \textit{reasoning beyond data}. For instance, in Scenario 4 in Figure \ref{fig:SHIFT} there is \textit{no} traffic light in the scene; however, the model is influenced by another deliberate question and predicts a red traffic light. Consequently, explanation systems, particularly large pre-trained models, must be constructed in a way that describes the visual scene as it is or just ``fails'' safely by not delivering an explanation, thereby avoiding the presentation of a potentially convincing but fictitious explanation. \\
\textit{3. Incorrect action, correct explanation}: Explaining an incorrect action of an AV correctly can help detect errors with the existing driving system. We relate this category to the fail-safe capability of such vehicles: When the AV detects its loss of/limited automation ability, it can safely stop at the side

\begin{figure*}[htp!]
 \centering
    \includegraphics[width=17 cm]{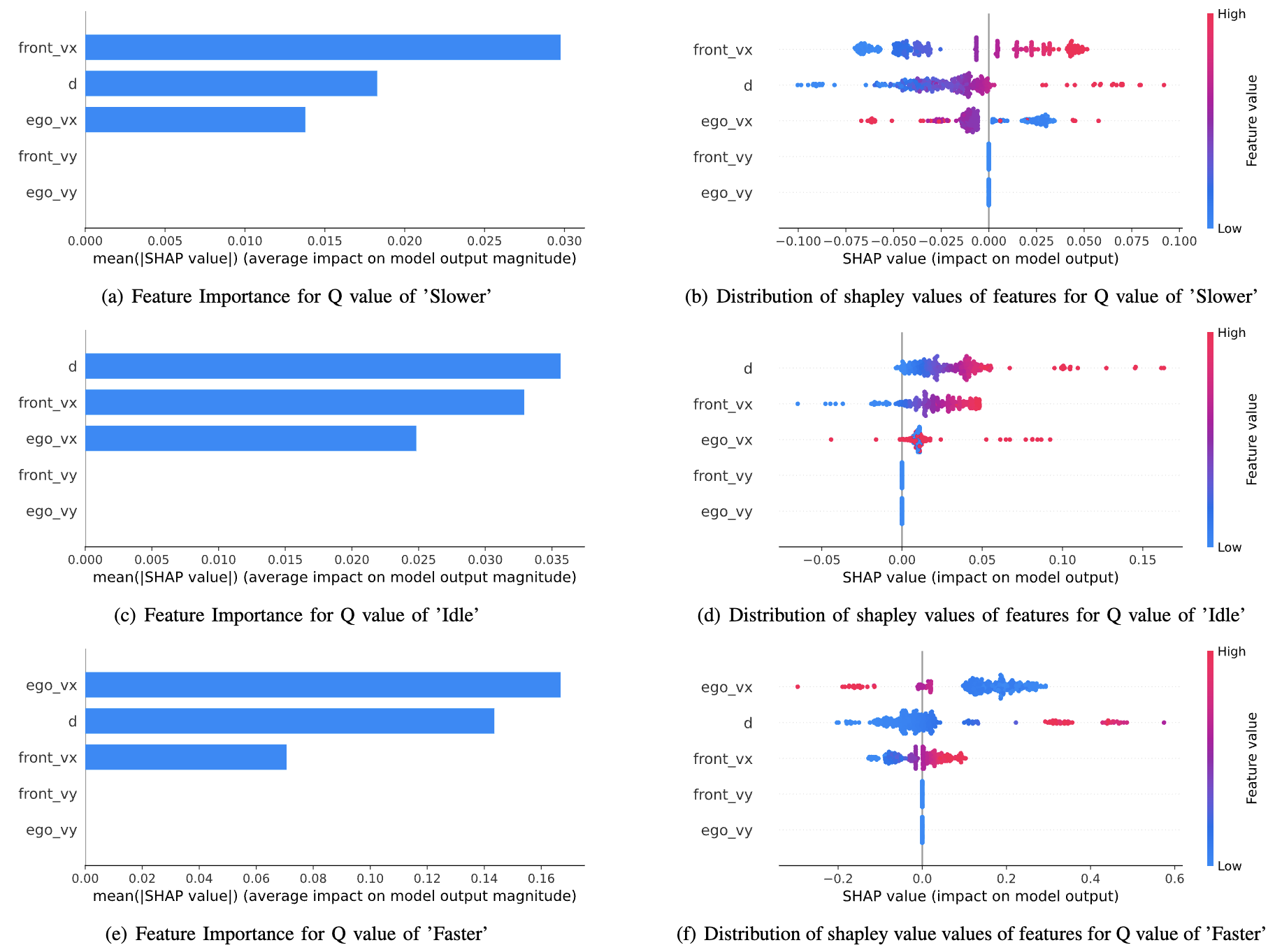}
    \caption{The results from Cui et al.'s \cite{cui2022interpretation} experiment on describing influential features on the action decisions of the AV via feature importance vs. average absolute Shapley value plots for actions ``Slower,"  ``Idle,"  and ``Faster," respectively, in the vehicle-following task.}
    \label{fig:shap_rl}
\end{figure*} \hspace{-0.45cm}
of the road temporarily and prevent potential hazards ahead. Likewise, when an AV does not stop at a red traffic light, for the question “Why did the AV continue to drive?” the explanation system can help with \textit{corrective} answer describing “It seems the AV made a mistake as it should have stopped as the red light was on.” This explanation can further help system engineers detect the issue with the perception of the AV.\\
\textit{4. Incorrect action, incorrect explanation}: Finally, in the fourth case, the unsafe behavior of an AV is accompanied by incorrect explanations. For instance, referring to Scenario 6 in Figure \ref{fig:SHIFT}, the AV keeps going under a \textit{red} light, and when a question is asked about its behavior, the explanation model generates an incorrect response describing the light turning \textit{green} (where it does not, actually). This is the most undesirable situation as the AV performs an unsafe action and the explanatory system cannot detect/describe it properly to carry out corrective measures. Such issues hinder the applicability of transformer-based QA models to be deployed as a reliable real-time explanation mechanism to HMIs of AVs.
\subsection{Experiment 2: Quantitative Analysis on Feature Importance Scores for Understanding Influential Driving Features}
Here, we describe a quantitative experimental analysis of an XAI method for enhancing the safety of end-to-end driving. We refer to the experimental study of Cui et al. \cite{cui2022interpretation}, in which the authors use the SHAP technique to interpret deep RL-based decisions of an AV in a vehicle-following task. In the experiment, the state space is represented by \textit{d} denoting the relative distance between the AV and the proceeding vehicle, $ego_{vx}$ denotes the longitudinal velocity of the AV, $ego_{vy}$ denotes the lateral velocity of the AV, $front_{vx}$ denotes the longitudinal velocity of the vehicle in front, and $front_{vy}$ denotes the longitudinal velocity of the leading vehicle:

\begin{equation}
    S=[d, ego_{vx}, ego_{vy}, front_{vx}, front_{vy} ]
\end{equation}

With this setup, the action space $A$ consists of three discrete action labels;  deceleration, keeping the current speed, and acceleration denoted with the following terms: 

\begin{equation}
    A=[Slower, Idle, Faster]
\end{equation}

Finally, three kinds of rewards are defined in terms of $r_d$ - driving as fast as possible, $r_v$ - keeping a safe distance from the proceeding vehicle, and $r_c$ - avoiding collision with the proceeding vehicle. The SHAP method is further employed to simplify the feature space and describe the dependency between features and the influence of each feature on an AV's action decision. Figure \ref{fig:shap_rl} shows the results of this experiment.
\subsection{Discussion}
\begin{figure*}[htp!]
 \centering
    \includegraphics[width=17.5 cm]{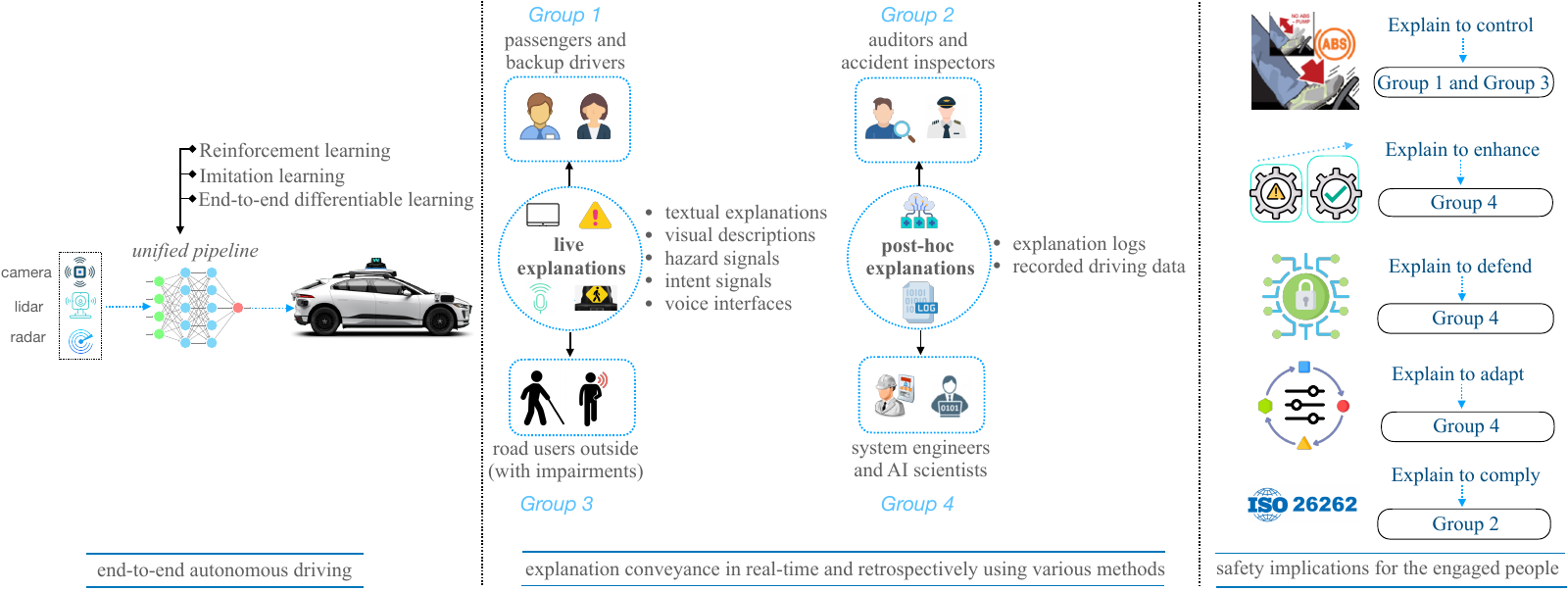}
    \caption{A high-level, illustrative diagram of safety implications of explanations for the engaged people in end-to-end autonomous driving}
    \label{fig:safety_summarization_fig}
\end{figure*}

The analysis of the results of the presented experiments discloses several findings from the safety perspective. Given that stakeholders of autonomous driving have various backgrounds, safety implications of relevant explanations are also different. In Experiment 1, the explanations are targeted at people on board an AV. While the correctness of responses helps them trust the action decisions of the vehicle while it operates on a road, inaccurate responses can trigger an alert and show the malfunction of the action-explanation mapping. Such awareness could make them pause the trip and prevent potential foreseeable danger ahead for themselves and people outside, particularly those with visual and hearing impairments. In addition, temporally evolving action-explanation log data delivered via human-interpretable language could help auditors and accident inspectors to deal with accountability and culpability issues appropriately in case of road mishaps with the presence of these vehicles. \\
On the contrary, quantitative analysis such as the one in Experiment 2 can help autonomous driving developers, such as system engineers and AI scientists, improve safety and efficiency of driving in several ways. As illustrated in Figure \ref{fig:shap_rl}, sub-figures (a), (c), and (e) show the importance of each feature with respect to mean
absolute Shapley value plots for the three discrete actions - ``Slower," ``Idle," and ``Faster." For each of the action types, $ego_{vx}$, $d$, and $front_{vx}$ are most influential, while the remaining features do not have a substantial effect on the actions of an AV. Hence, the results suggest that  $ego_{vy}$, and $front_{vy}$ can be removed from state representation. Such retrospective analysis could have several safety benefits for the driving system. First, feature importance scores can help adjust the reward function to shape the learning process of the RL model and prevent potential reward misdesign issues in autonomous driving \cite{knox2023reward}. Furthermore, the model complexity can also be reduced via such a quantitative post-hoc analysis. Eventually, properly defined feature space, state-space representation, and reward formulation can help cope with scalable evaluation \cite{guo2019safe, zhou2024scalable}, uncertainty quantification \cite{michelmore2020uncertainty}, runtime monitoring \cite{grieser2020assuring} of end-to-end driving. \\
Overall, based on the broad spectrum of analytical and empirical analyses throughout our study, we can measure the value of explanations from the safety perspective at least in five ways: \\
\textit{Explain to control}: Real-time or live explanations may enable human drivers/passengers on board to intervene and control critical scenarios that lead to effective human-machine teaming and a safer trip. Clear and concise explanations can reduce the cognitive load on back-up drivers and enable quicker responses during critical moments. Furthermore, a proper combination of visual, auditory, and haptic feedback could help back-up drivers receive and comprehend takeover requests promptly. \\
\textit{Explain to enhance}: Retrospective explanations may help detect system errors and sensor malfunctions, enabling the implementation of corrective measures that reduce the risk of further accidents caused by the faulty behavior of an AV. By analyzing explanations of past decisions as log data, system developers can identify errors or weak points in the system and improve the end-to-end pipeline. In addition, patterns observed via post-hoc explanations can guide software updates, ensuring continuous improvement of the driving system.\\ 
\textit{Explain to defend}: Explanations can assist in identifying vulnerabilities and security flaws and developing robust cybersecurity measures to prevent malicious attacks on the behavior of an AV, as described in Experiment 1. Such information can reveal patterns in how adversarial attacks exploit weaknesses in the driving system, helping developers address these vulnerabilities. Moreover, by analyzing those patterns, explanations can guide the creation of adversarial examples for training, improving the driving system's ability to handle such inputs.\\
\textit{Explain to adapt}: Explanations can facilitate continuous improvement. By understanding how the deployed driving system responds to various (especially rare and unseen) situations, system developers and engineers can refine the system through adaptive learning and enhance safety features over time. Particularly, by explaining decisions in previously unseen driving scenarios, the system can adapt to new environments and incorporate these learnings into future actions and help with learning to handle such situations more effectively in the future.\\
\textit{Explain to comply}: AI transparency can ensure that a self-driving system's decisions align with established regulations, making it easier for authorities to assess the overall safety of end-to-end driving and approve at least semi-AVs for widespread use by society.  The values explanations can create are demonstrating accountability, facilitating audits, supporting accident investigation, aligning with ethical standards,  and aiding certification/licensing processes, which are crucial principles of automotive regulatory compliance. \\
To summarize, safety implications of explanations for relevant interaction partners in end-to-end driving are described in Figure \ref{fig:safety_summarization_fig} with representative illustrations. While these safety benefits also apply to modular or any autonomous driving architectures, it is noteworthy to underscore that end-to-end driving is a monolithic system directly mapping sensor information to control commands via RL, IL, or differentiable learning. On the other hand, modular autonomous driving is a decomposed system consisting of several interconnected modules, where modular systems need explainability analysis both within individual modules (such as perception and planning modules) and across module interactions. A promising analysis of the safety implications of XAI in modular autonomous driving has been recently performed by the SafeX framework in \cite{kuznietsov2024explainable}, as previously noted, and we recommend readers refer to this framework for more insights.
\section{Limitations and Future Directions}
The sections above summarize the safety implications of XAI approaches in end-to-end driving analytically and empirically via scenario-based evaluations within a specific scope. However, explainability-informed safety assessment of end-to-end autonomous driving goes beyond the scope of our study. For instance, detecting explanation anomalies, robustness and time constraints of explanations, and explanations in uncertainty with confidence-explanation correlation and uncertainty

\hspace{-0.40cm}
trigger rates, along with their safety guarantees, have not been studied in our paper. These nuances could be considered the limitations of our study. Meanwhile, we leave these limitations as a motivation for future explanation and shed light on the potential safety ramifications of these aspects briefly for end-to-end driving.
\subsection{Building Robust Interactive Explanations}
Building robust interactive explanation models for AVs remains yet another challenging task. The experiment described in Section VI.A discloses that even advanced interactive explanation frameworks, particularly transformer-based QA models, are prone to present falsified responses to human adversarial questions.
Meanwhile, various studies, such as \cite{kim2023and, mohseni2021multidisciplinary}, reveal that incorrect explanations could have a negative impact on users’ perceived safety and mental model on an autonomous system. As transformed-based QA models are increasingly deployed on end-to-end AVs as a dialogue method \cite{wayve_team}, such issues may seriously damage users’ perceived safety of an AV and negatively affect regulatory approval of this technology. \\
A potential solution could be incorporating human-adversarial examples into the training process for driving specific QA models. For instance, when a passenger onboard
asks the HMI, ``Why is the vehicle turning to the left?” as an adversarial question in an actual right turn scenario under a green light, the HMI could present an action-reflecting
response like ``No, the vehicle is turning to the right as the traffic light allows a right turn,” explaining AV action with causal attribution. Regardless of whether such
questions are asked intentionally or unintentionally, the explanation models should understand the joint semantics of question and scene and present context-aware responses.
Consequently, pre-trained models must be regulatable by construction \cite{kenny2023pursuit}, ensuring their inference process leverages \textit{human-defined concepts}. Thus, interactive explanations in general could also be designed to support counterfactual reasoning and neuro-symbolic reasoning, which combines deep perception with symbolic logic to explain traffic rules, intentions, and causal events. In this regard, an intelligent driving system must be robust against any potential perturbations while conveying interactive explanations.

\subsection{Reaching a Consensus on Time Constraint of Explanation Conveyance}
The timing mechanism of explanations remains under-explored compared to the form and content of explanations. As autonomous driving decisions are time-critical from a safety perspective, it is of paramount importance to investigate the timing mechanism within two essential directions: 1) Lead time from automated to human-controlled mode, 2) Considering people with various functional and cognitive abilities need explanations so that these messages become helpful and not result in information overload for them. As end-to-end driving is considerably data-hungry and requires processing a huge amount of data in real time, the relevant HMIs must deliver explanations on time in an adequate format. This feature is vital for takeover cases and passengers' real-time situation awareness and safety monitoring of self-driving decisions. \\ 
Viable approaches to an effective timing mechanism could be a reached consensus between automotive manufacturers \cite{schieben2019designing} and investigating people's diversity, satisfaction with, and reaction to specific modes of explanation in real time \cite{shin2024enhancing}. For instance, extensive scenario-based evaluations with human actors in the loop can help understand the time needed for takeover situations. While the mentioned directions can help meet the users' explanation demands effectively, they also have significant implications for real-time driving safety and legal culpability aspects in case of post-accident investigations.  
\vspace{-0.5cm}
\subsection{Safety Analysis with Explanations in Uncertainty}
Despite substantial advancements in providing a variety of explanations for autonomous driving decisions, conveying the level of confidence to relevant interaction partners in such explanations remains a significant challenge. This issue deepens in uncertainty in the driving environment, such as in adverse weather conditions and situations with reduced
visibility (e.g., dense fog, nighttime driving), necessitating uncertainty-aware safe perception. Without carefully measuring residual risks and environmental uncertainty, overconfident decisions may have dire consequences for an AV and human actors at that traffic scene. Consequently, uncertainty estimation is a vital problem in dealing with unforeseen events safely and having situation awareness with uncertain information \cite{endsley1995toward, sanneman2022situation, endsley2023supporting}.
While quantifying uncertainty has recently been investigated by autonomous driving researchers from several aspects, such as for statistical guarantees \cite{michelmore2020uncertainty}, and object detection \cite{meyer2020learning, peng2021uncertainty}, providing relevant explanations with a certain level of confidence under uncertainty currently remains relatively unexplored in end-to-end autonomous driving. To our knowledge, only a few studies have attempted to explore explainability within uncertainty, such as \cite{pan2020towards} and \cite{ling2024improving}, and the literature remains scarce with relevant studies in general. Hence, as dynamic driving environments often involve considerable uncertainties, there is an imminent need to justify AV actions with an appropriate degree of confidence. Instead of solely presenting deterministic explanations to people, granular explanations can help understand residual risks in scenarios with various hazard levels.

\section{Conclusion}
In this paper, we have presented an investigation of safety implications of XAI approaches in end-to-end autonomous driving. Through critical case studies and empirical evidence, our paper reveals the value of explanations in enhancing end-to-end driving safety and shows the potential advantages, limitations, and challenges of explanations for achieving this goal. Our study considers informational content and form, the timing perspective of explanations, and their conveyance to targeted interaction partners for in-depth safety analysis. We can provide two main conclusions based on the spectrum of our study: 1) Explanations can help reveal \textit{latent patterns} in training data of an end-to-end model that led to failures and traffic accidents as a data-driving insight, and 2) Such failure and more generally, situation awareness explanations, could guide relevant data augmentation techniques to enhance the end-to-end model’s behavior via \textit{iterative learning}, ultimately enhancing driving safety in ways that the modular approach might miss.\\ 
Thus, to our knowledge, this paper is the first to explore safety benefits and consequences of XAI techniques in end-to-end driving from multidimensional aspects. We trust that the presented guidelines can help improve the safety of end-to-end driving and build trustworthy, transparent, and publicly acceptable AVs in the new era of autonomous driving.

\section*{Acknowledgment}

We acknowledge support from the Alberta Machine Intelligence Institute (Amii), from the Computing Science Department of the University of Alberta, and the Natural Sciences and Engineering Research Council of Canada (NSERC).

\ifCLASSOPTIONcaptionsoff
  \newpage
\fi



%

\bibliography{IEEEabrv.bib, main.bib}{}
\bibliographystyle{IEEEtran}

%
\vspace{-1cm}
\begin{IEEEbiography}
    [{\includegraphics[width=1in,height=1.25in,clip, keepaspectratio]{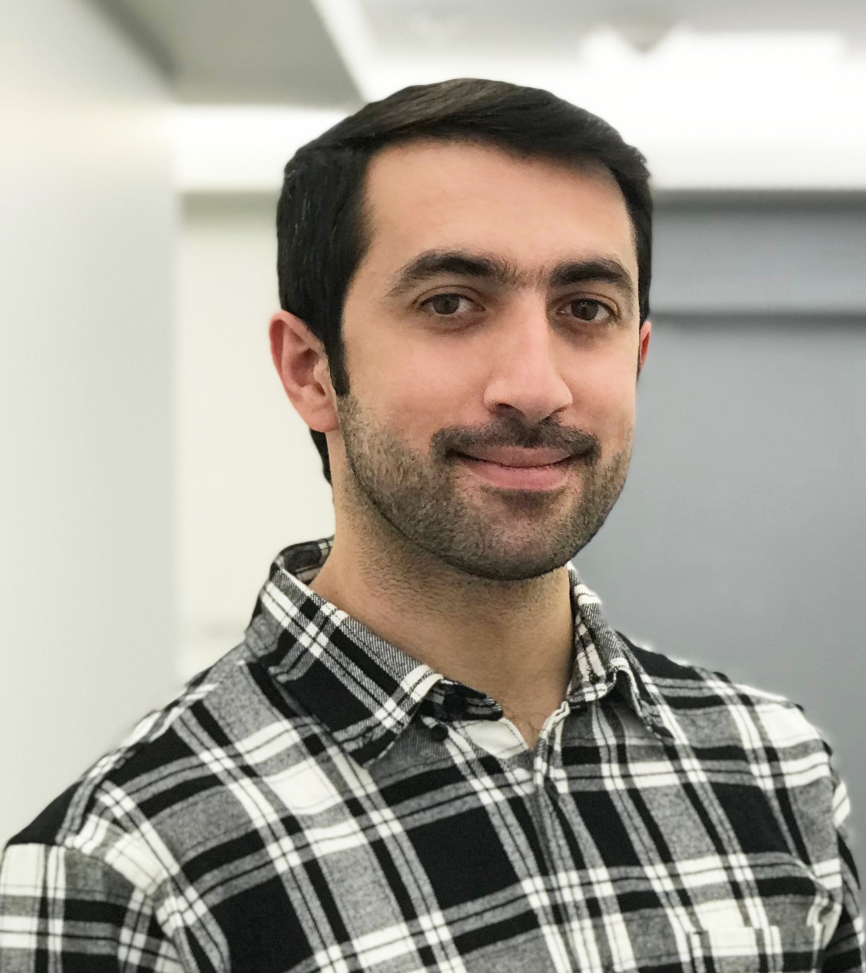}}]{Shahin Atakishiyev}
is from the Mirzabayli village of Gabala, Azerbaijan. He received a PhD in Computing Science in September 2024 and an MSc in Software Engineering and Intelligent Systems in January 2018 from the University of Alberta, Canada. He obtained a BSc in Computer Engineering from Qafqaz University, Azerbaijan, in June 2015. During his doctoral study, Shahin built explainable artificial intelligence (XAI) approaches for autonomous vehicles under the supervision of Prof. Randy Goebel. His research interests include safe, ethical, human-centered, and explainable artificial intelligence applied to real-world problems.
\end{IEEEbiography}

\begin{IEEEbiography}
    [{\includegraphics[width=1in,clip, keepaspectratio]{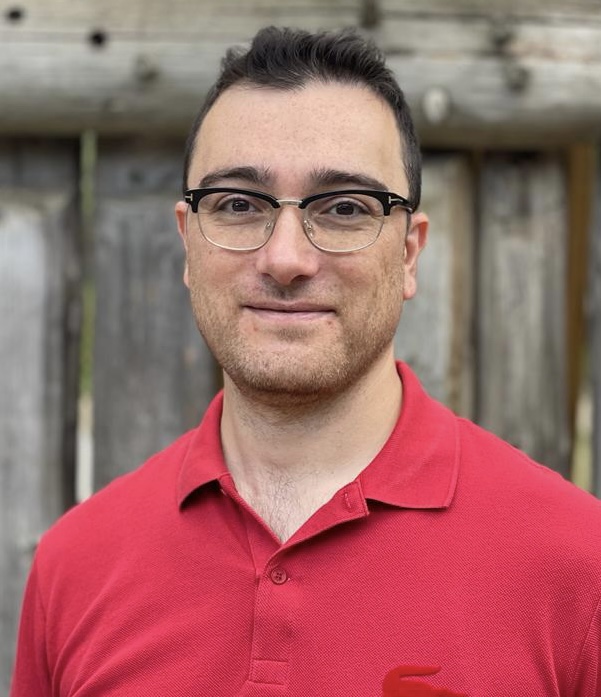}}]{Mohammad Salameh}
received the Ph.D. degree from the University of Alberta under the supervision of Dr. Greg Kondrak and Dr. Colin Cherry, with the main focus on statistical machine translation and sentiment analysis. He is currently a Principal Researcher at Huawei Technologies Canada Company Ltd and leading the neural architecture search group, focusing on gradient-based and reinforcement learning approaches. He co-organized Determining Sentiment Intensity in Tweets (SemEval2016) and Affects in Tweets (SemEval2018) shared tasks.
\end{IEEEbiography}

\begin{IEEEbiography}
[{\includegraphics[width=1in,clip, keepaspectratio]{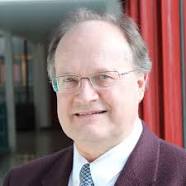}}]{R.G. (Randy) Goebel}
 is currently a Professor of Computing Science in the Department of Computing Science at the University of Alberta and Fellow and co-founder of the Alberta Machine Intelligence Institute (Amii). He received the B.Sc. (Computer Science), M.Sc. (Computing Science), and Ph.D. (Computer Science) from the Universities of Regina, Alberta, and British Columbia, respectively.

Professor Goebel's theoretical work on abduction, hypothetical reasoning and belief revision is internationally well known; his recent research is focused on the formalization of visualization and explainable artificial intelligence (XAI), especially in applications in autonomous driving, legal reasoning, and precision health. He has worked on optimization, algorithm complexity, systems biology, natural language processing, and automated reasoning.

Randy has previously held faculty appointments at the University of Waterloo, University of Tokyo, Multimedia University (Kuala Lumpur), Hokkaido University (Sapporo), visiting researcher engagements at National Institute of Informatics (Tokyo), DFKI (Germany), and NICTA (now Data61, Australia); he is actively involved in collaborative research projects in Canada, Japan, Germany, France, the UK, and China.
\end{IEEEbiography}




\end{document}